%% file: main.tex
\documentclass[lettersize,journal]{IEEEtran}
\usepackage{amsmath,amsfonts}
\usepackage{algorithmic}
\usepackage{algorithm}
\usepackage{array}
\usepackage[caption=false,font=normalsize,labelfont=sf,textfont=sf]{subfig}
\usepackage{textcomp}
\usepackage{stfloats}
\usepackage{url}
\usepackage{verbatim}
\usepackage{graphicx}
\usepackage{cite}
\usepackage{color}
\usepackage[inkscapelatex=false]{svg}
\usepackage{booktabs}
\usepackage{multirow}
\usepackage{multicol}
\usepackage{verbatim}
\usepackage{makecell}
\usepackage{amsmath}
\usepackage{threeparttable}
\usepackage{lscape}
\usepackage[figuresright]{rotating}

\begin{document}
\setlength{\belowcaptionskip}{-1cm} 
\title{Towards Few-Shot Learning in the Open World: \\
A Review and Beyond}

\author{Hui~Xue$^*$,~\IEEEmembership{Member,~IEEE},Yuexuan~An,Yongchun~Qin,Wenqian Li,Yixin~Wu,Yongjuan~Che,
		Pengfei~Fang, and~Min-Ling Zhang,~\IEEEmembership{Senior~Member,~IEEE}
		\IEEEcompsocitemizethanks{\IEEEcompsocthanksitem  H. Xue, Y. An, Y. Qin, W. Li, Y. Wu, Y. Che and P. Fang are with the School of Computer Science and Engineering, Southeast University, Nanjing 210096, China and the Key Laboratory of New Generation Artificial Intelligence Technology and Its Interdisciplinary Applications (Southeast University), Ministry of Education, Nanjing 211189, China. M. Zhang is with the School of Computer Science and Engineering, Southeast University, Nanjing 210096, China and the Key Laboratory of Computer Network and Information Integration (Southeast University), Ministry of Education, Nanjing 211189, China.
			\protect\\
			E-mail: \{ hxue, yx\_an, ycqin, wenqianli.li, wuyx, yjche, fangpengfei, zhangml\}@seu.edu.cn
		}
		\thanks{$^*$ Corresponding author}
}

\maketitle

\begin{abstract}

Human intelligence is characterized by our ability to absorb and apply knowledge from the world around us, especially in rapidly acquiring new concepts from minimal examples, underpinned by prior knowledge. Few-shot learning (FSL) aims to mimic this capacity by enabling significant generalizations and transferability. However, traditional FSL frameworks often rely on assumptions of clean, complete, and static data, conditions that are seldom met in real-world environments. Such assumptions falter in the inherently uncertain, incomplete, and dynamic contexts of the open world. This paper presents a comprehensive review of recent advancements designed to adapt FSL for use in open-world settings. We categorize existing methods into three distinct types of open-world few-shot learning: those involving varying instances, varying classes, and varying distributions. Each category is discussed in terms of its specific challenges and methods, as well as its strengths and weaknesses. We standardize experimental settings and metric benchmarks across scenarios, and provide a comparative analysis of the performance of various methods. In conclusion, we outline potential future research directions for this evolving field. It is our hope that this review will catalyze further development of effective solutions to these complex challenges, thereby advancing the field of artificial intelligence.

\end{abstract}


\input{chapters/introduction}

\input{chapters/OFSL}
\input{chapters/Noisy}

\input{chapters/Open-set}
\input{chapters/Incremental}
\input{chapters/cross-domain/Cross-domain}

\input{chapters/Performances}

\input{chapters/cross-domain/performance_CD}

\input{chapters/Open_Challenges_and_Future_Detections}

\section{Conclusion}

The open world poses a critical challenge in the realm of artificial intelligence, spanning applications such as self-driving cars, medical analysis, and remote sensing. Open-world few-shot learning (OFSL) emerges as a generalization of few-shot learning (FSL), aiming to enhance FSL performance within the diverse and dynamic open-world context. The core of OFSL lies in the capacity to detect and adapt to new, varied, and ever-changing scenarios. This paper delineates OFSL settings, encompassing varying instances, classes, and distributions. We comprehensively review the evolution of related methods and conduct comparative analyses of representative approaches within each OFSL scenario. Additionally, we outline several challenging open problems that merit substantial research endeavors and efforts in the future.

\section*{Acknowledgments}
This work was supported by the National Natural Science Foundation of China (Nos.62076062 and 62306070) and the Social Development Science and Technology Project of Jiangsu Province (No.BE2022811). Furthermore, the work was also supported by the Big Data Computing Center of Southeast University.


%

\bibliographystyle{IEEEtran}

\bibliography{reference}

 \vspace{-15mm}

\begin{IEEEbiography}[{\includegraphics[width=0.8in,height=1in]{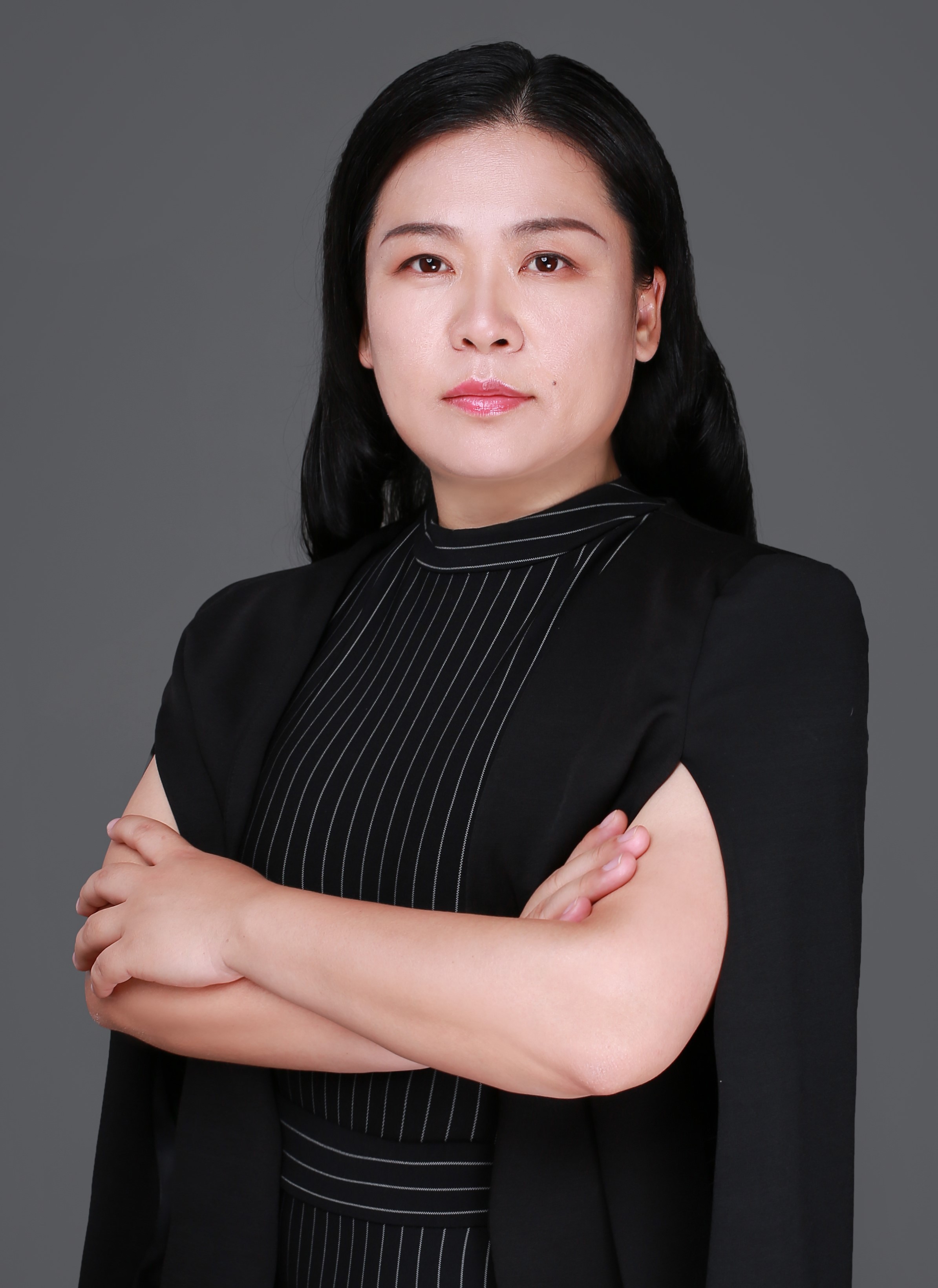}}]{Hui Xue}(Member, IEEE)
    \scriptsize
		is currently a professor of School of Computer Science and Engineering at Southeast University, China. She received the B.Sc. degree in Mathematics from Nanjing Norm University in 2002. In 2005, she received the M.Sc. degree in Mathematics from Nanjing University of Aeronautics \& Astronautics (NUAA). And she also received the Ph.D. degree in Computer Application Technology at NUAA in 2008. Her research interests include pattern recognition and machine learning.
	\end{IEEEbiography}

 \vspace{-10mm}

\begin{IEEEbiography}[{\includegraphics[width=0.8in,height=1in]{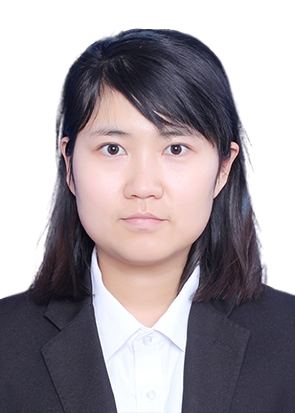}}]{Yuexuan An}
    \scriptsize
		received the B.Sc. in computer science and technology from Jiangsu Normal University in 2015 and M.Sc. degree in Computer Application Technology in China University of Mining and Technology in 2019. She is currently pursuing the Ph.D. degree in School of Computer Science and Engineering, Southeast University. Her research interest includes machine learning and pattern recognition.
	\end{IEEEbiography}

 \vspace{-10mm}

\begin{IEEEbiography}[{\includegraphics[width=0.8in,height=1in]{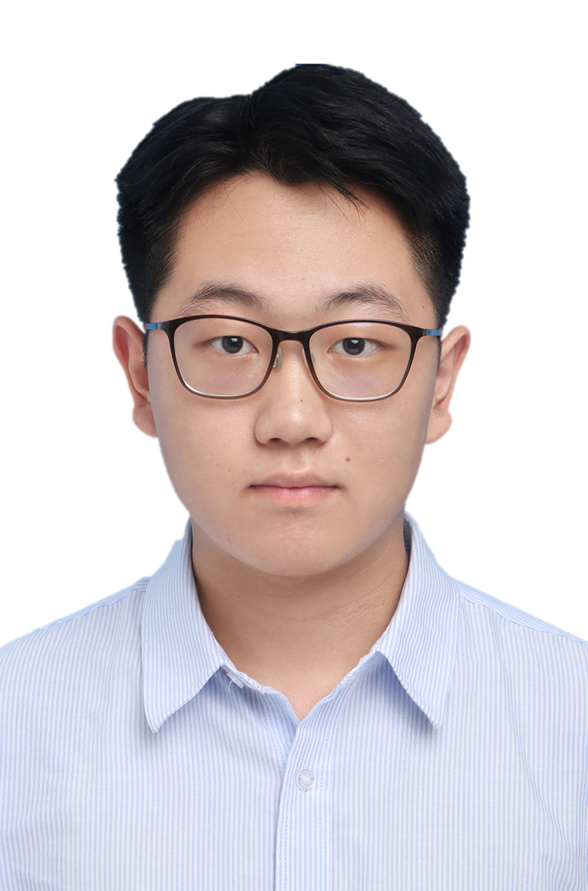}}]{Yongchun Qin}
    \scriptsize
		received the B.Sc. in artificial intelligence from Southeast University in 2023. He is currently pursuing the M.Sc. degree in School of computer science and engineering, at Southeast University. His research interest includes machine learning and pattern recognition.
	\end{IEEEbiography}

 \vspace{-10mm}

 \begin{IEEEbiography}[{\includegraphics[width=0.8in,height=1in]{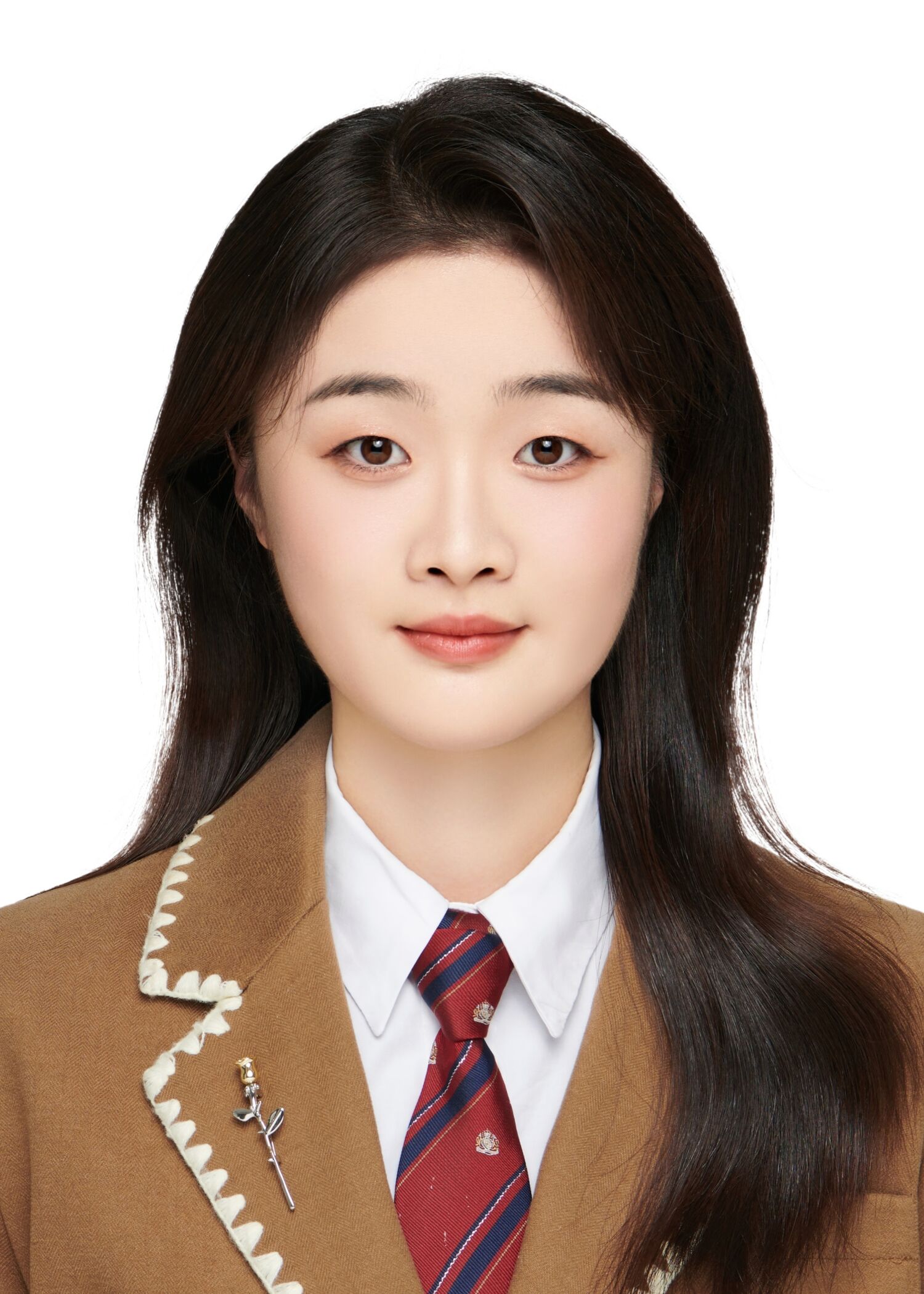}}]{Wenqian Li}
     \scriptsize
        received the B.E. in information science and engineering technology from Southeast University in 2023. She is currently pursuing the Ph.D. degree in School of computer science and engineering, Southeast University. Her research interest includes machine learning and pattern recognition.
    \end{IEEEbiography}

 \vspace{-10mm}

\begin{IEEEbiography}[{\includegraphics[width=0.8in,height=1in, clip,keepaspectratio]{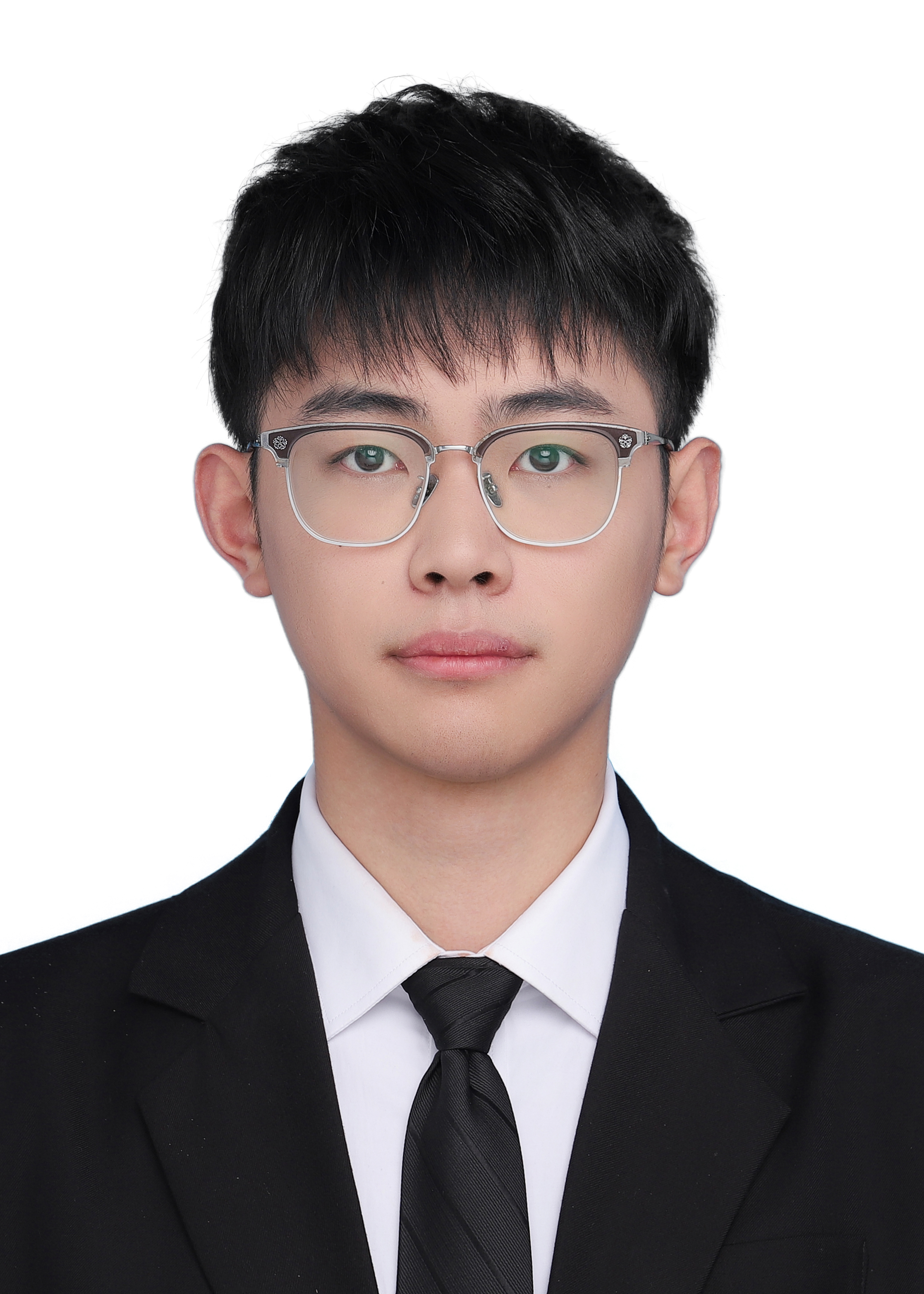}}]{Yixin Wu} 
    \scriptsize
    received the B.Sc. in computer science and technology from Nanjing University of Information Science \& Technology in 2022. He is currently pursuing the M.Sc. degree in School of computer science and engineering, Southeast University. His research interest includes machine learning and pattern recognition. \end{IEEEbiography}

 \vspace{-10mm}

\begin{IEEEbiography}[{\includegraphics[width=0.8in,height=1in]{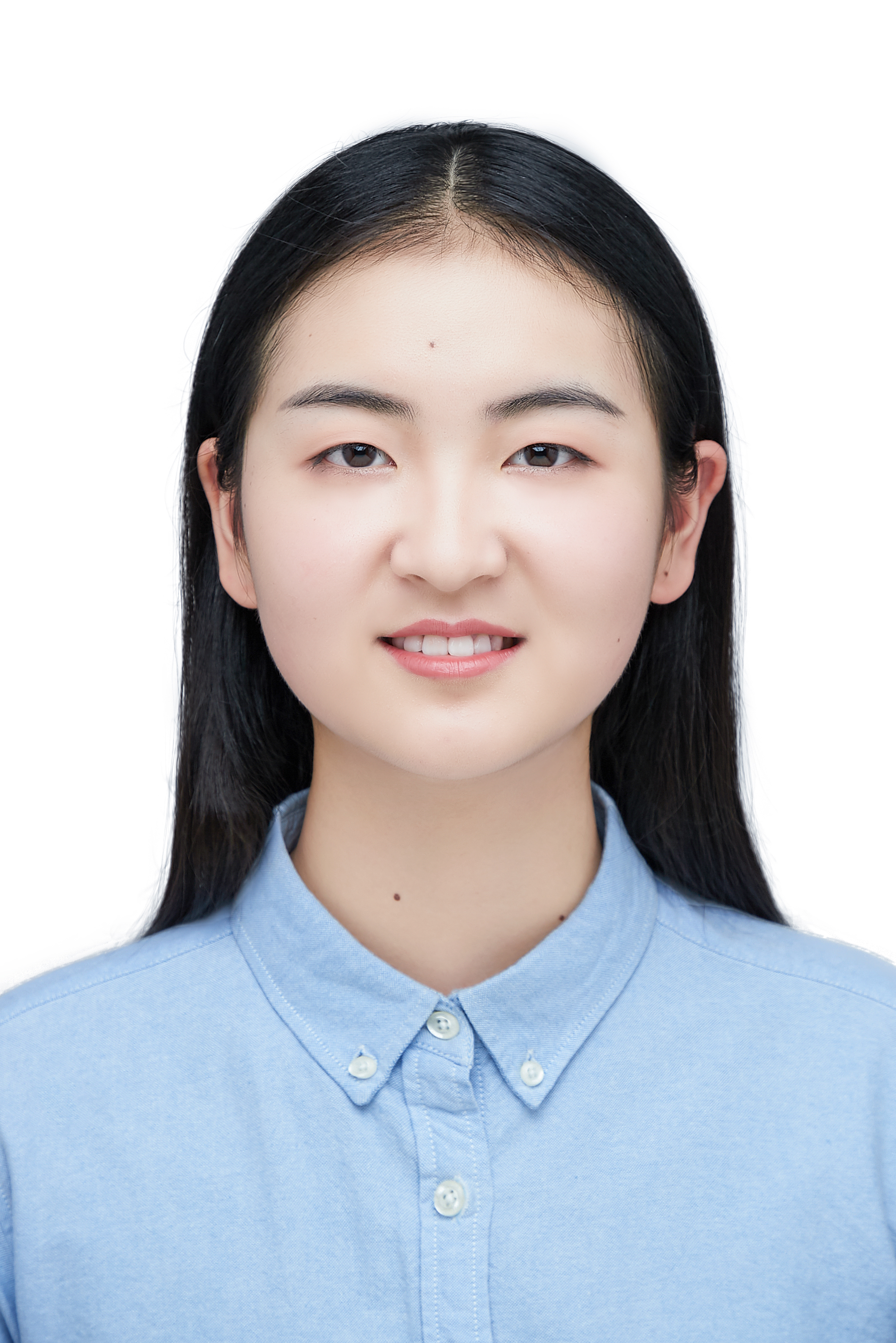}}]{Yongjuan Che}
    \scriptsize
        received the B.Sc. in Computer Science and Technology from Hohai University in 2021 and M.Sc. degree in Software Engineering from Southeast University in 2024.  Her research interest includes machine learning and pattern recognition.
    \end{IEEEbiography}

 \vspace{-10mm}

	\begin{IEEEbiography}[{\includegraphics[width=0.8in,height=1in]{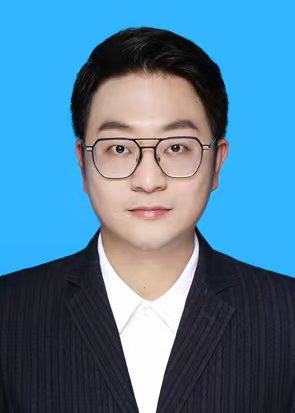}}]{Pengfei Fang}
     \scriptsize
		is an Associate Professor at the School of Computer Science and Engineering, Southeast University (SEU), China. Before joining SEU, he was a post-doctoral fellow at Monash University in 2022. He received the Ph.D. degree from the Australian National University and DATA61-CSIRO in 2022, and the M.E. degree from the Australian National University in 2017. His research interests include computer vision and machine learning.
	\end{IEEEbiography}

 \vspace{-10mm}

		\begin{IEEEbiography}[{\includegraphics[width=0.8in,height=1in]{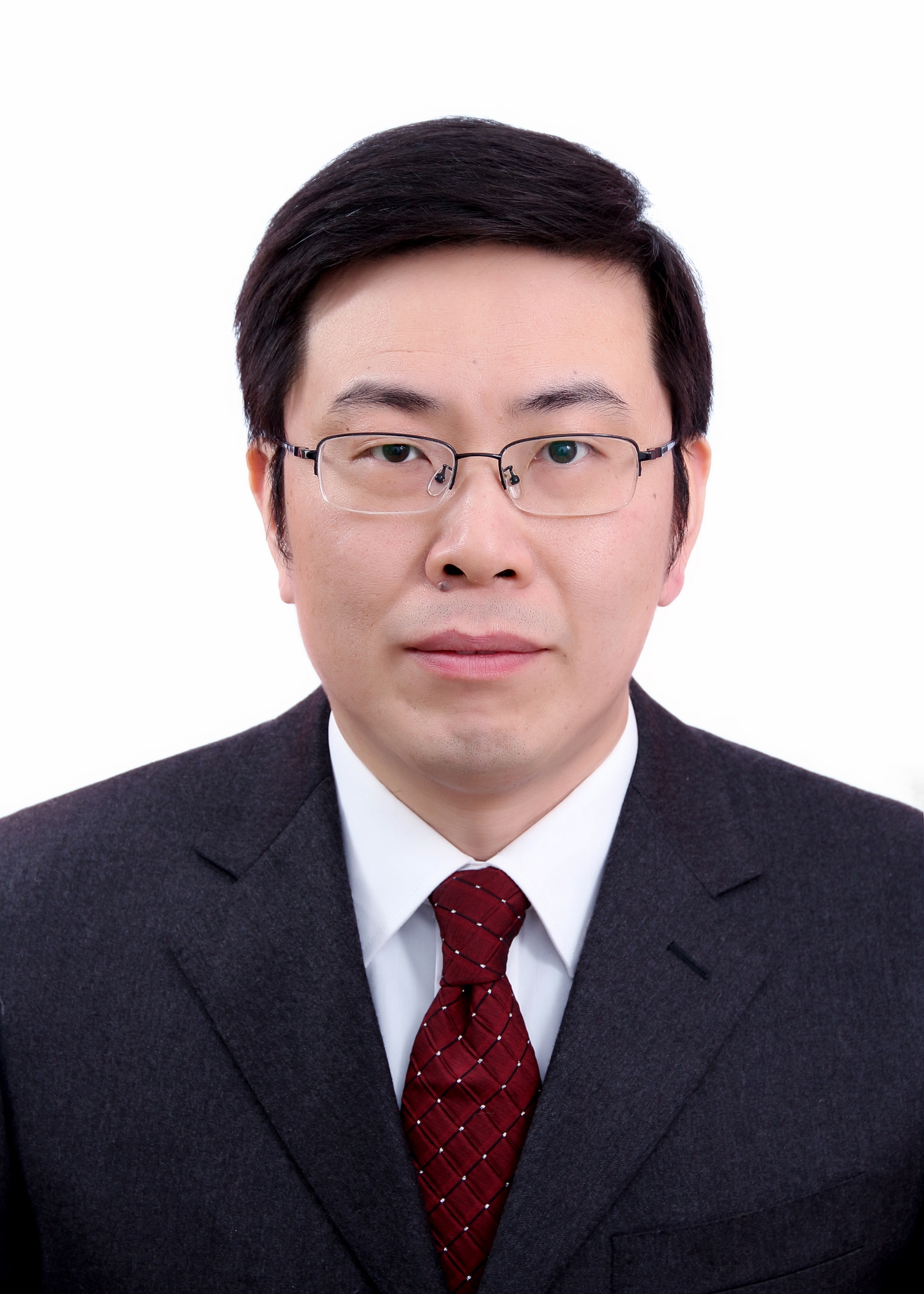}}]{Min-Ling Zhang}(Senior Member, IEEE)
    \scriptsize
		received the BSc, MSc, and PhD degrees in computer science from Nanjing University, China, in 2001, 2004 and 2007, respectively. Currently, he is a Professor at the School of Computer Science and Engineering, Southeast University, China. His main research interests include machine learning and data mining. In recent years, Dr. Zhang has served as the General Co-Chairs of ACML’18, Program Co-Chairs of CCML’25, PAKDD’19, CCF-ICAI’19, ACML’17, CCFAI’17, PRICAI’16, Senior PC member or Area Chair of KDD 2021-2024, AAAI 2022-2025, IJCAI 2017-2024, ICML 2024, ICLR 2024, etc. He is also on the editorial board of IEEE Transactions on Pattern Analysis and Machine Intelligence, Science China Information Sciences, ACM Transactions on Intelligent Systems and Technology, Frontiers of Computer Science, Machine Intelligence Research, etc. Dr. Zhang is the Steering Committee Member of ACML and PAKDD, Vice-Chair of the CAAI (Chinese Association of Artificial Intelligence) Machine Learning Society. He is a Distinguished Member of CCF, CAAI, and Senior Member of AAAI, ACM, IEEE.
	\end{IEEEbiography}

\vfill

\end{document}

%% file: chapters/Introduction.tex
\section{Introduction}

\IEEEPARstart{I}{n} recent years, the explosive growth of annotated data coupled with the rapid advancements in device computing capabilities has propelled deep learning to exhibit remarkable resilience, significantly advancing the field of artificial intelligence. However, these achievements hinge on the availability of vast amounts of labeled training data to construct predictive models endowed with robust generalization capabilities. Nevertheless, in many practical application scenarios, limitations such as manpower, resources, and environmental constraints often hinder the acquisition of a sufficient number of labeled training samples. This limitation severely compromises the effectiveness of machine learning models in scenarios with scarce labeled data. In contrast, human intelligence demonstrates the ability to swiftly establish cognition of new concepts with minimal learning samples, often requiring only one or a few instances \cite{lake2015human}. 

To simulate this powerful learning ability of humans, Few-Shot Learning (FSL) has emerged as an exceptional machine learning paradigm, which learns to recognize new concepts from just a few examples\cite{FSL2,FSL3,Closer}. FSL excels in scenarios without relying on very expensive and time-consuming data collection \cite{TransMatch}. Despite its effectiveness, traditional FSL algorithms often operate under the strict assumption that all data observations must be described by a fixed, complete, and clean data space. This assumption necessitates uniformity across instances, classes, and distributions. However, in various practical worlds, their conditions cannot always be met, making FSL a more challenging task. 

To wit, consider medical image analysis aided by FSL, in which the collection of scarce medical data is a continuous and laborious process involving contributions from clinical experts from different regions. In this circumstance, challenges abound low-quality annotations of ambiguous images serving as examples, unobserved disease types to be predicted, constantly emerging new subspecies of diseases necessitating model updates, and the prevalence of non-in-distribution images within training sets. Instead of being fixed and known as advanced, the data space used to describe the FSL data in such applications is varying and dynamic, which should not be ignored in the open world.

\begin{figure*}
	\centering
	\includegraphics[width=1.95\columnwidth]{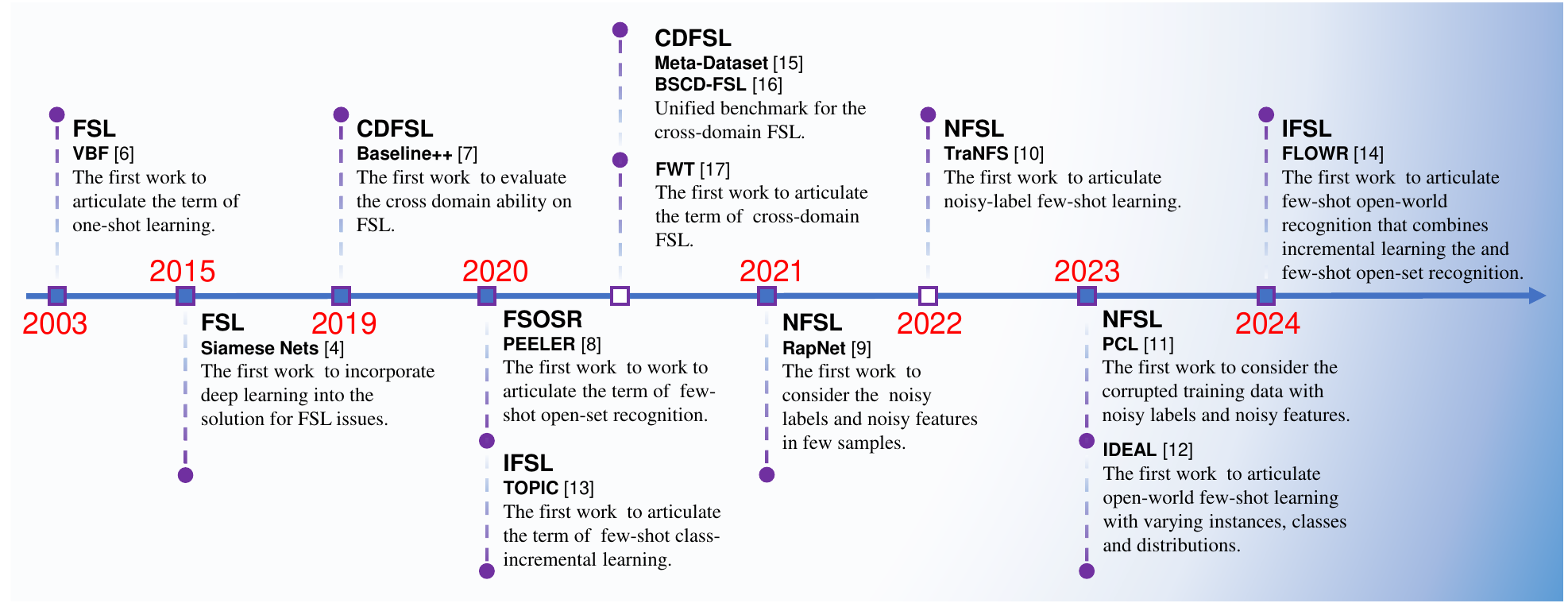}
	\caption{Chronological milestones on FSL and OFSL, including representative FSL and OFSL approaches and the related benchmarks. FSL was first noticed as a topic in 2003 \cite{LiF2003}. Subsequently, it underwent continuous developments \cite{SiameseNet,Closer}. After that, different open-world few-shot learning scenarios have been proposed successively, including few-shot learning with varying instances \cite{PEELER,RFSL, TraNFL,PCL,IDEAL}, few-shot learning with varying classes \cite{TOPIC,WillesJ2024} and few-shot learning with varying distributions \cite{c76_triantafillou2019meta,c75_guo2020broader,c28_tseng2020cross} in the open world.
	}
	\label{fig:introduction}
 \vspace{-3mm}
\end{figure*}

Recently, we have witnessed a surge of studies that aim at addressing the challenge of FSL in the open world, leading to the emergence of variants such as cross-domain FSL (CDFSL)\cite{Closer}, incremental FSL (IFSL)\cite{TOPIC}, open-set FSL (a.k.a. Few-Shot Open-Set Recognition, FSOSR)\cite{PEELER}, noisy FSL (NFSL)\cite{TraNFL}, and more, as shown in Fig. \ref{fig:introduction}. These variants are regarded as general cases of FSL tasks confronted with the varying and dynamic nature of the open world. They enable the machine to learn in a manner akin to human learning,  thereby bridging the gap between human cognition and machine learning. Several existing surveys have made detailed summaries and prospects for FSL, especially in some specific settings such as cross-domain FSL and incremental FSL. In studies \cite{FSL3,LiWYDTQHSWGL23,LuGYZZ23}, authors delve into the significant challenges of FSL and review numerous FSL methodologies within closed-world contexts.  However, beyond the fixed settings of the closed world, attention has also turned towards some specific open scenarios. In studies \cite{abs-2303-08557,abs-2303-09253}, researchers focus on cross-domain FSL, which addresses the performance degradation induced by domain gaps arising from the varying distributions between the source domain and the target domain in the open world.  Additionally, in \cite{TianLLRNT24,abs-2308-06764}, the incremental FSL has been explored, which addresses the catastrophic forgetting of previous knowledge and adaptability to the constant emergence of new classes in FSL tasks within the open world. 

However, the existing studies on one specific problem have been divided into multiple communities. Each community possesses its own set of scenario assumptions and design intuitions, leading to disparate and confusing evaluations of model performances. This parallel evolution of ideas has made communication increasingly difficult and complex, resulting in a lack of systematic analysis and reviews available for FSL in the open world.  

To address this gap and foster future research while providing newcomers with a comprehensive understanding of this challenging problem, this paper presents, for the first time, a comprehensive summary and review of FSL in the open world. Necessitated by the status quo of division, we deliver a timely review of prior arts and, more importantly, strive to unify and contextualize them under an umbrella paradigm, termed Open-world Few-Shot Learning (OFSL). We relax the traditional FSL assumption of a clean, complete, static data space, to an incomplete and dynamic nature of the data space encountered in the open world. We taxonomize existing OFSL into three main categories, including few-shot learning with varying instances, few-shot learning with varying classes and few-shot learning with varying distribution in the open world. Each scenario covers various aspects ranging from their related settings and methods, strengths and weaknesses, dataset, evaluation metrics and performance comparisons of representative methods. Furthermore, we describe their relatively few-shot scenarios and highlight this emerging topic's challenges and potential future directions.

\textbf{Specific contributions of our survey are as follows:}
\begin{itemize}
	\item We present a timely and the first survey of recent few-shot learning studies that transitioned from closed feature spaces to open feature spaces. By framing and analyzing the prior arts, our review provides a unified perspective that can bridge the gap between fragmented research communities and foster communication.
	
	\item We introduce a novel taxonomy that classifies existing researches into three prevailing open-world scenarios including few-shot learning with varying instances, few-shot learning with varying classes and few-shot learning with varying distribution in the open world. This taxonomy provides a useful framework for understanding the current state of the field and identifying areas for further research.
	
	\item We conduct an in-depth illustration of the representative methods in each scenario, analyzing their strengths and weaknesses, the common datasets and evaluation metrics, and comparing the performances of related methods. Our analysis provides valuable insights into the effectiveness of different approaches and the challenges involved in each scenario.
	
	\item We identify the open challenges confronted by existing few-shot learning studies and endeavor to shed light on untrodden pathways for future research. Our goal is to stimulate further exploration and discovery in this field to help researchers and practitioners navigate the complexities of few-shot learning in open-world scenarios.
\end{itemize}

%% file: chapters/OFSL.tex
\section{Problem Definition}

In this section, we provide the definition of few-shot learning and classify different scenarios in open-world few-shot learning with detailed explanations for them.

\vspace{-3mm}
\subsection{Few-shot Learning}
For $N$-way $K$-shot problems, we are given two datasets: a training set with a few labeled samples which is called support set $\mathcal{S}=\{({{\textbf{\textit{x}}}_{i}},{{y}_{i}})\}_{i=1}^{N\times K}$ and a test set consisting of unlabeled samples which is called query set $\mathcal{Q}=\{({{\textbf{\textit{x}}}_{i}},{{y}_{i}})\}_{i=NK+1}^{NK+M}$. Among them, ${\textbf{\textit{x}}}_{i}$ denotes the $i$-th sample, $y_i\in \mathcal{C}_{novel} $ from novel class is its corresponding label, $N$ is the number of classes in $\mathcal{S}$, $K$ is the number of samples extracted for each class, and $M$ is the number of samples in $\mathcal{Q}$. Each FSL problem that estimates the class of samples in the query set with the support set can be regarded as a task (episode). Meanwhile, an auxiliary dataset with abundant labeled samples which is called base class datasets $\mathcal{D}_{base} = \{({{\textbf{\textit{x}}}_{i}},{{y}_{i}})\}_{i=1}^{I}$. $\mathcal{D}_{base}$ is also divided into support sets $\mathcal{S}_{base}$ and query sets $\mathcal{Q}_{base}$ is used for meta-training, where ${y}_{i}\in \mathcal{C}_{base}$ and $\mathcal{C}_{base} \cap \mathcal{C}_{novel} {=}\emptyset$. This strategy can be regarded as a rehearsal to train a model on the data from the base classes $C_{base}$ so that the model can generalize well to the novel classes $C_{novel}$.

\vspace{-3mm}
\subsection{Open-world Few-Shot Learning}
Open-world Few-Shot Learning (OFSL) considers the scenarios in which the FSL datasets emerge unseen concepts that may appear in $\mathcal{D}_{base}$, $\mathcal{S}$ or $\mathcal{Q}$. According to the different ways in which unknown concepts exist, OFSL can be divided into three categories, including few-shot learning with varying instances in the open world, few-shot learning with varying classes in the open world and few-shot learning with varying distributions in the open world.

\subsubsection{Few-Shot Learning with Varying Instances in the Open World} 
Few-shot learning with varying instances in the open world considers two typical problems, including noisy few-shot learning and few-shot open-set recognition. Noisy few-shot learning considers the scenario that the samples in the training set are no longer unreliable and some samples may be polluted by noise. The noisy samples may exist in $\mathcal{D}_{base}$ or $\mathcal{S}$, and the noise may come from both feature space and label space. few-shot open-set recognition focuses on the scenario that the query set $\mathcal{Q}$ includes unknown class queries which can be denoted as $\mathcal{C}_{unknown}$. That is, a query sample $\left( \boldsymbol{x}_{q}, {y}_{q}  \right)$ belongs to either $\mathcal{C}_{novel}$ or $\mathcal{C}_{unknown}$. Therefore, the objective of few-shot open-set recognition is to predict the class label of the query accurately if ${y}_{q} \in \mathcal{C}_{novel}$ or to identify the unknown query sample if ${y}_{q} \in \mathcal{C}_{unknown}$.

\subsubsection{Few-Shot Learning with Varying Classes in the Open World} 
Few-shot learning with varying classes in the open world considers the incremental few-shot learning problem. It supposes there is a stream of labeled training sets $\mathcal{D}_{1}, \mathcal{D}_{2}, \cdots$. The FSL model is incrementally trained $\mathcal{D}_{1}, \mathcal{D}_{2}, \cdots$ while only $\mathcal{D}_{t}$ is available at the $t$-th training session. The classes added in different sessions do not have any overlap, i.e., $\forall i,j$ where $i\ne j$, ${\mathcal{C}_{i}} \cap {\mathcal{C}_{j}}{=}\emptyset$. After the incremental session $T$, the model is evaluated on test data belonging to the classes seen in both the current session and all previous sessions, i.e., $\mathcal{C}_{1} \cap \mathcal{C}_{2}, \cdots \cap \mathcal{C}_{T}$.

\subsubsection{Few-Shot Learning with Varying Distributions in the Open World} 
Few-shot learning with varying classes in the open world considers the cross-domain few-shot learning problem. It aims at the scenario that varying degrees of significant domain gaps exist between source and target domains. Considering a source domain $\mathcal{D}_{sourse} =\left(\mathcal{X}_{sourse}, \mathcal{Y}_{sourse}  \right)$ (usually regarded as the base class datasets $\mathcal{D}_{base}$) with joint distribution $\mathcal{P}_{{sourse}}$ and a target domain $\mathcal{D}_{target} = \left(\mathcal{X}_{target}, \mathcal{Y}_{target} \right)$ (usually regarded as the novel class datasets $\mathcal{D}_{novel}$) with joint distribution $\mathcal{P}_{target}$, the goal of cross-domain few-shot learning is to transfer prior knowledge from the source domain $\mathcal{D}_{sourse}$ to the target domain $\mathcal{D}_{target}$, where $\mathcal{D}_{sourse} \neq \mathcal{D}_{target}$, and $\mathcal{P}_{sourse} \neq \mathcal{P}_{target}$.

%% file: chapters/Noisy.tex
\section{Few-Shot Leaning with Varying Instances in the Open World: Noisy Few-Shot Learning}
\begin{figure*}[t]
\centering
    \includegraphics[width=1.7\columnwidth]{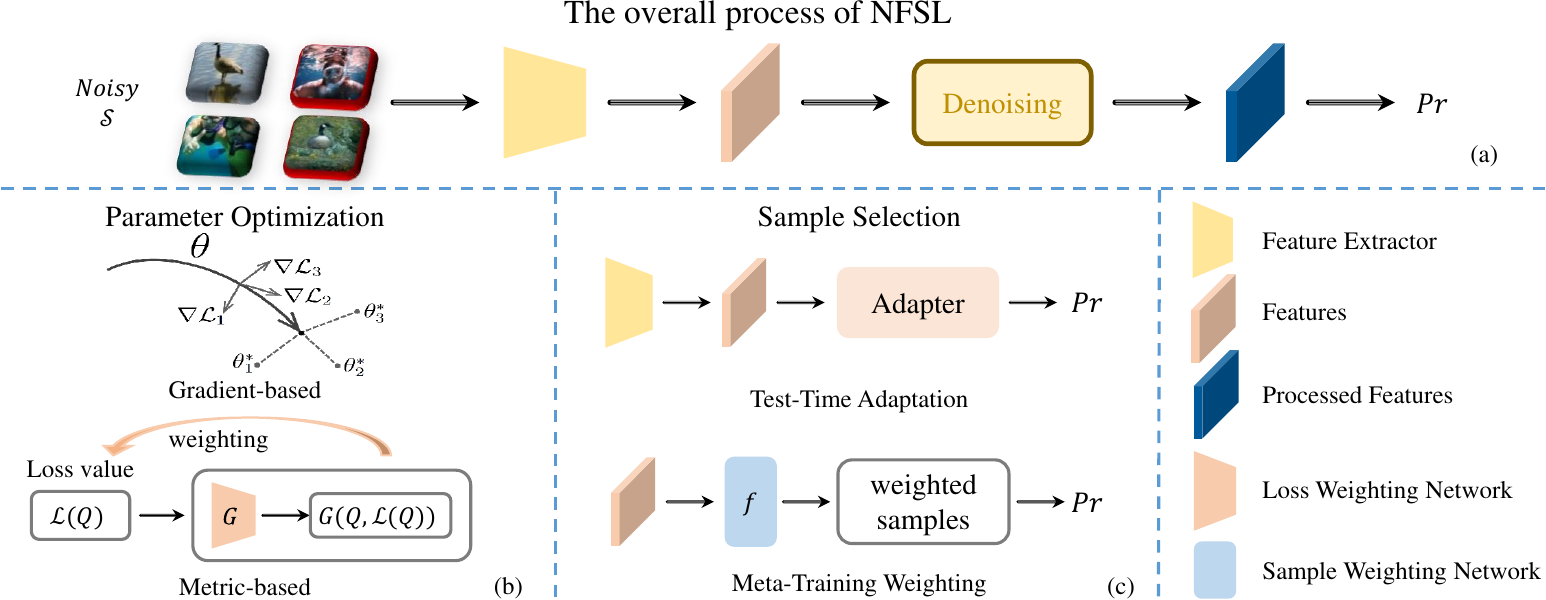}
    \caption{An overall diagram of existing NFSL researches. (a) Overview of the NFSL setting.(b) Parameter Optimization methods are mainly divided into Gradient-based and Metric-based methods. (c)The sample selection methods are mainly divided into Test-Time Adaptation and Meta-Training Weighting.}
    \label{fig:nfsl}
     \vspace{-2mm}
\end{figure*}

In the open-world environment, few-shot learning systems often face the challenge that instances vary frequently. The varying instances can be divided into two situations based on the range of instances, i.e., noisy instances and unknown instances: noisy few-shot learning and few-shot open set recognition. The problem of few-shot learning with noisy instances, also known as noisy few-shot learning, requires maintaining performance when faced with noisy samples in the support set. The key challenges of noisy few-shot learning come from two aspects: 1) insufficient data:  due to the rarity of samples, the potential noise is hard to detect with insufficient data, and 2) uncertainty information: each sample might contain important category information, incorrect recognition of noise and training can lead to devastating uncertainty on model training. Therefore, FSL models should eliminate the uncertainty and disturbance from noisy samples. To address this issue, several works have been devoted to adapting to the noisy environment in FSL. We categorize existing methods into two classes: parameter optimization and sample selection, as shown in Fig. \ref{fig:nfsl}. The methods based on parameter optimization are devoted to optimizing the model by assigning proper weights to the losses of the tasks or the instance. Common approaches include gradient-based weighting (e.g. NESTED MAML \cite{NESTEDMAML}, Eigen-Reptile \cite{Eigen-Reptile}) and metric-based weighting (e.g. PCL \cite{PCL}). The methods based on sample selection aim to assign lower weight to the sample which is considered as noise. Common approaches include test-time adaptation (e.g. RNNP \cite{RNNP}, DETA \cite{DETA}, CO3 \cite{CO3}) and meta-training weighting (e.g. RapNets \cite{RFSL}, TraNFS \cite{TraNFL}, IDEAL \cite{IDEAL}).

\begin{figure}[t]
\centering
    \includegraphics[width=0.95\columnwidth]{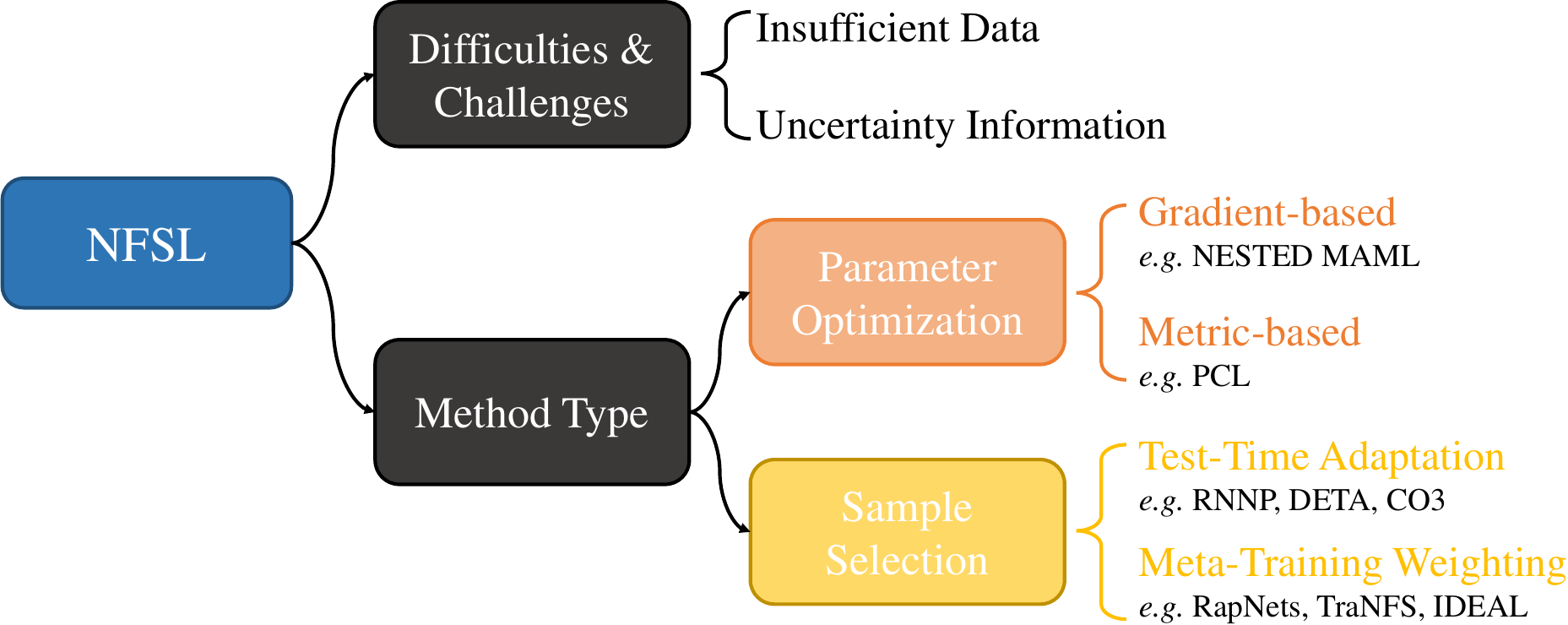}
    \caption{The structural relationships of the existing two types of methods and their subclasses, as well as the difficulties and challenges faced by NFSL problems.}
    \label{nfsl}
     \vspace{-2mm}
\end{figure}

In this section, we will introduce the latest progress of few-shot learning with varying instances from two aspects. The overview of the problem definition and method strategy is depicted in Fig. \ref{nfsl}.

\vspace{-3mm}

 \vspace{-3mm}
\subsection{Parameter Optimization}
Parameter optimization learns the optimal weights of the losses or the tasks and the parameter update direction to improve the model performance under noisy scenarios. According to the way of parameter update, the parameter optimization can further be divided into gradient-based methods and metric-based methods. 

Gradient-based methods, as depicted in Fig. \ref{nfsl}(b), fine-tune model parameters on a few support examples. NESTED MAML \cite{NESTEDMAML} modifies MAML by assigning weights to the losses of both tasks and solving the weight in a nested bi-level optimization. It studies the general form of task and instance-weighted meta-learning, where the optimal weights and model initialization parameters are learned by optimizing a nested bi-level objective function. Eigen-Reptile \cite{Eigen-Reptile} constructs prior models to decide which sample should be discarded with self-paced
learning to learn more accurate main direction. The main direction is computed efficiently, with the scale of the calculated matrix related to the number of gradient steps rather than the number of parameters.

Metric-based methods classify samples based on their similarities. PCL \cite{PCL} adopts a collaborative mechanism to reweight the loss according to the global ranks of losses and local intra-class correlation information. This mechanism extracts the sample importance information from both global and local aspects by encoding the feature representations and the corresponding loss values of the training samples.

\noindent \textbf{Summary and Discussion.}

Parameter optimization does not require additional data preprocessing and can be applied to various types of models, including deep learning models and traditional machine learning models.
However, these methods depend too much on the choice of weighting strategy that inappropriate strategy may lead to performance degradation. Besides, they are not sensitive to hard samples and outliers which could be assigned higher weights, adversely affecting model training.

\subsection{Sample Selection}
Sample Selection, as depicted in Fig. \ref{nfsl}(c), learns the weights of the few-shot examples to alleviate the impact of the noisy samples. According to the way of denoising, the sample selection can further be divided into test-time adaptation methods and meta-training weighting methods. 

Test-time adaptation methods mainly focus on dealing with noises in base classes, rather than in the few-shot task. RNNP \cite{RNNP} adopts the Mixup \cite{Mixup} strategy to synthetic samples with samples with the same label and cluster all features to generate new classes. It generates mixed features of samples and adopts k-means to produce refined prototypes.
DETA \cite{DETA} is a meta-training-free method, that focuses on filtering out task-irrelevant, noisy representations. It crops each image into different regions to extract more relevance in images and weights each region according to the cosine similarity of paired regions to filter image noise and label noise. CO3 \cite{CO3} effectively leverages the potential and prior knowledge of foundation models to enhance model performance. It introduces TeFu-Adapter, a specially designed mechanism that reduces the negative impact of noisy labels during the adapter updating stage.

Meta-training weight methods mainly focus on dealing with the noise in the current few-shot task. Dr.k-NN \cite{Dr.k-NN} modifies the classifier with a weighted k-NN classifier by weighting k neighbor samples. It first resorts the weighted k-NN problems to finite-dimensional convex programming. Then, it solves the convex optimization by jointly learning the feature embedding and the minimax optimal classifier. RapNets \cite{RFSL} regenerates sample features in each class according to the similarity between the paired samples within a class and encodes sample features to BiLSTM to achieve the weight of each sample. It proposed an attentive module to learn the weight of samples in the same class. TraNFS \cite{TraNFL} achieves the weight of each sample according to the attention value in transformer by concatenating samples as a sequence to filter label noise. It leverages a modified version of the self-attention mechanism to downweight samples considered to be mislabeled. IDEAL \cite{IDEAL} proposes a unified framework to implement comprehensive calibration from instance to metric. It leverages dual networks, where one extracts intra-class similarity and the other extracts inter-class divergence, to jointly weight samples. IDEAL reevaluates each sample by aggregating the information of dual networks while considering the consistent constraints. DCML \cite{DCML} proposes a curriculum meta-learning model with a dual-level class-example sampling strategy that formulates a robust curriculum to adaptively adjust the task distribution for robust model training.

\noindent \textbf{Summary and Discussion.}

Sample selection can delicately remove noisy samples, which easily enhances the model's generalization to unseen data. However, these methods involve additional data preprocessing steps, including identification and removal of noisy labels, which may increase implementation complexity. Besides, sample selection may introduce bias about how to select and identify noisy samples.

%% file: chapters/Open-set.tex
\section{Few-Shot Leaning with Varying Instances in the Open World:  Few-Shot Open-Set
Recognition}

\begin{figure}[t]
\centering
    \includegraphics[width=0.95\columnwidth]{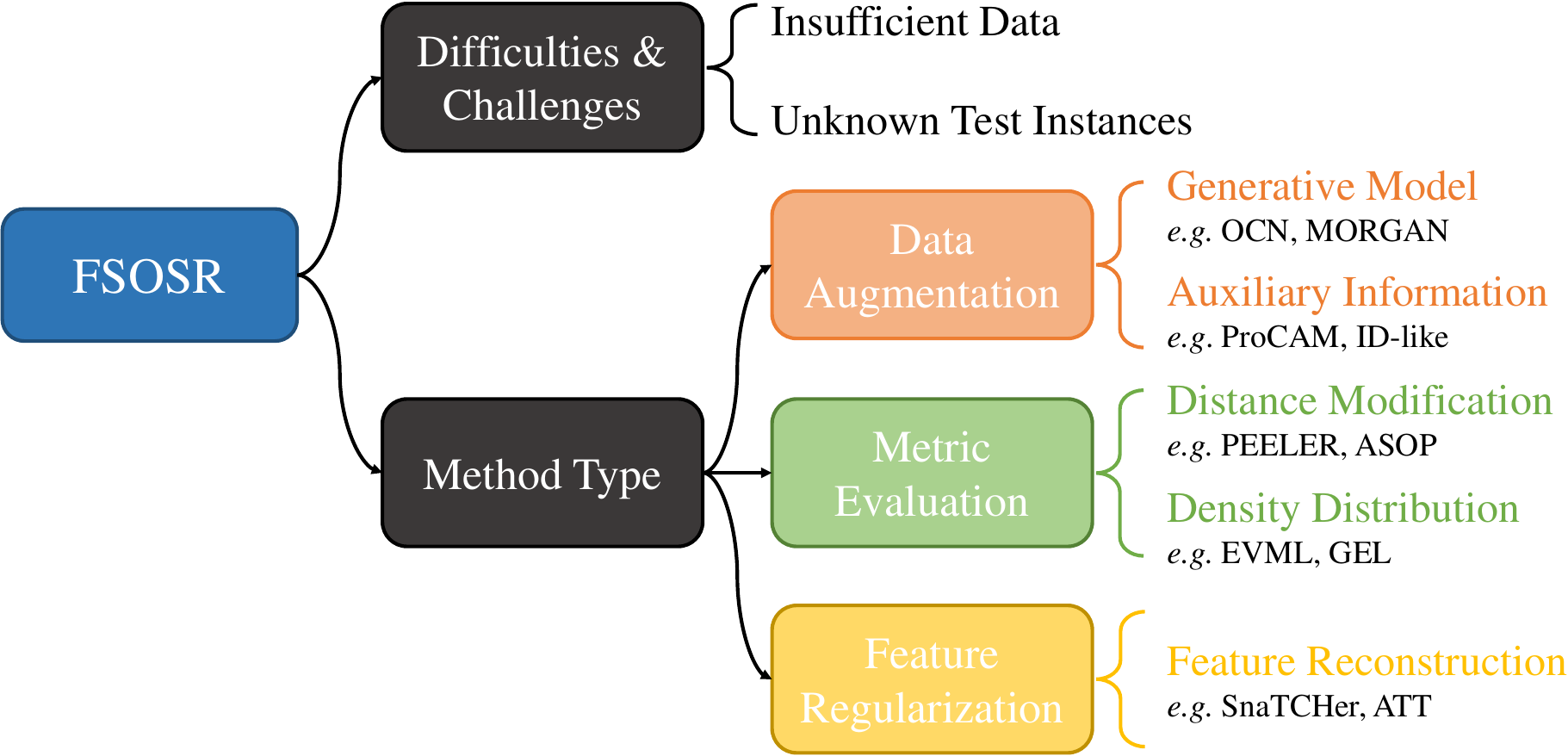}
    \caption{The structural relationships of the existing three types of methods and their subclasses, as well as the difficulties and challenges faced by FSOSR problems.}
     \label{fig:opensetFrame}
      \vspace{-2mm}
\end{figure}

In the open-world environment, few-shot systems often face the challenge of varying test samples that do not belong to known classes. The problem of few-shot learning with unknown test instances, also known as few-shot open-set recognition (FSOSR), requires maintaining classification performance on known classes and rejecting the unknown ones. The key challenges of FSOSR come from two aspects: 1) insufficient data: due to the scarcity of samples, it is difficult to construct precise decision boundaries for known classes and 2) unknown instances: the model needs to detect unknown test instances while simultaneously classifying known classes. Therefore, FSL models should fully capture the distribution information of known classes with limited samples and learn the true distribution of unknown classes to simulate the test environment. To address this issue, several works have focused on adapting to the open-set environment, which can be categorized into a unified taxonomy including Data Augmentation, Metric Evaluation, and Feature Processing, as shown in Fig. \ref{fig:opensetFrame}. The data-based approaches generate pseudo samples to simulate the distribution of unknown classes. Common approaches include generative models(e.g.OCN\cite{OutlierCalibrationNetwork}, MORGAN
\cite{palMORGANMetaLearningBasedFewShot2023} and DA-FSOS\cite{palDomainAdaptiveFewShot2023}) and auxiliary information (e.g. ProCAM\cite{songFewshotOpensetRecognition2022} and ID-like
prompts\cite{ID-like}). The metric-based approaches construct a highly discriminative feature space. The mainstream technological paths include distance modification (e.g. PEELER\cite{PEELER}, FSOSTC\cite{liFewShotOpenSetTraffic2022}, openFEAT\cite{kishanOpenFEATImprovingSpeaker2022},
CTR\cite{dionelisCTRContrastiveTraining2022}, MRM \cite{cheBoostingFewShotOpenSet2023}, 
ASOP\cite{kimTaskAgnosticOpenSetPrototype2023},
MSCL \cite{zhao2024meta} and 
G-FOOD \cite{su2024toward}) and density distribution (e.g.EVML\cite{palExtremeValueMetaLearning2023a}, OSLO \cite{boudiafOpenSetLikelihoodMaximization2023a} and
GEL\cite{wangGlocalEnergybasedLearning2023}). Feature-based methods compare the differences between the reconstructed sets before and after (e.g. SnaTCHer \cite{SnaTCHer},
ATT\cite{huangTaskAdaptiveNegativeEnvision2022},
ReFOCS \cite{nagReconstructionGuidedMetaLearning2023a},
FreqDiMFT\cite{FreqDiMFT} and 
RFDNet\cite{dengLearningRelativeFeature2023}).

\begin{figure*}[t]
\centering
    \includegraphics[width=1.7\columnwidth]{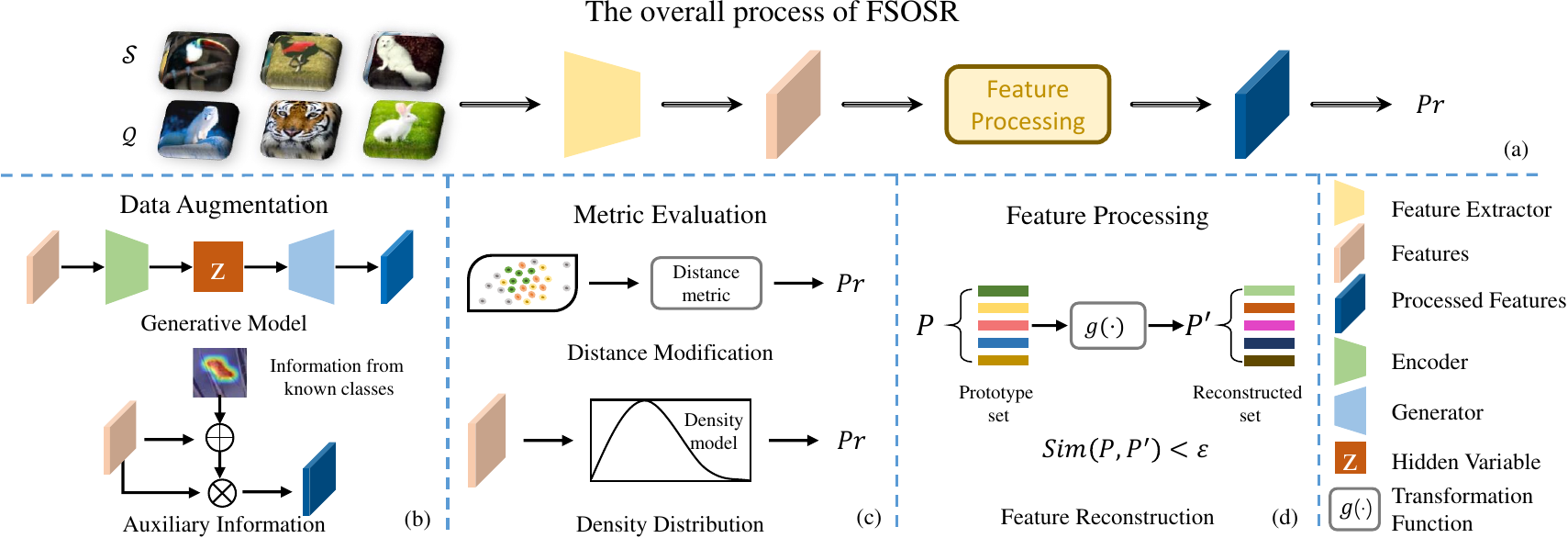}
    \caption{An overall diagram of existing FSOSR researches. (a) Overview of the FSOSR setting.(b) Data Augmentation methods are mainly divided into Generative Model and Auxiliary Information. (c) Metric Evaluation methods include Distance Modification and Density Distribution. (d)Feature Processing methods compare the differences between the reconstructed sets before and after using transformation functions.}
   \label{fig:opensetExample}
    \vspace{-3mm}
\end{figure*}

In this section, we will introduce the latest progress of few-shot learning with unknown instances from three aspects. The overview of the problem definition and method strategy is depicted in Fig. \ref{fig:opensetExample}.

\vspace{-3mm}
\subsection{Data Augmentation} 
Data augmentation, as depicted in Fig. \ref{fig:opensetExample}(b), enriches the original training set to tackle the issue of insufficient labeled training samples. According to the source of pseudo samples, data augmentation can further be divided into generative model methods and auxiliary information methods.

Generative model methods generate pseudo samples using probabilistic statistical models or machine learning models. OCN\cite{OutlierCalibrationNetwork} proposes using samples generated by a Variational Autoencoder (VAE) to augment the query set features, applying it to hyperspectral image recognition. It generates pseudo-known samples by optimizing the VAE using the known class support set and query sets, allowing it to recognize samples generated by the VAE as known classes and detecting samples from unknown classes. Building on this, MORGAN \cite{palMORGANMetaLearningBasedFewShot2023} combines FSOSR with Generative Adversarial Networks (GAN) and uses two different noise variances to generate pseudo-known and pseudo-unknown samples, including a decaying Gaussian distribution with lower noise variance to generate pseudo-known samples and a generator with higher noise variance to simulate features of unknown classes. DA-FSOS\cite{palDomainAdaptiveFewShot2023} involves augmenting both known and pseudo-unknown classes from source and target domains using a pair of class and domain conditional adversarial networks. 

Generative models trained with limited samples may produce low-quality pseudo samples, resulting in a decline in model performance. To solve this problem,  auxiliary information is adopted in FSOSR. ProCAM \cite{songFewshotOpensetRecognition2022} suggests using the background information of the support set as the unknown class, proposing that the background of support samples significantly differs from the foreground known class which can be known as an extra background unknown class. ID-like
prompts\cite{ID-like} leverages the CLIP model to construct ID-like outliers from ID samples through random cropping and filtering based on cosine similarity with ID prompts. The adaptive prompts are learned by aligning OOD prompts with outliers and maximizing the dissimilarity between prompts to improve classification probabilities. 

\noindent\textbf{Summary and Discussion.} Data augmentation methods can directly tackle the challenges of limited training samples and the unpredictable nature of unknown classes by utilizing pseudo-samples, which enhances the model's ability to generalize unknown classes.  However, pseudo-unknown samples generated might not accurately represent the true distribution of unknown classes. Besides, due to the small size of the support set, the amount of auxiliary information it can provide is limited.

\subsection{Evaluation Metric } 
Metric evaluation, as depicted in Fig. \ref{fig:opensetExample}(c), constructs an appropriate feature space and evaluates the similarity between samples using a metric function. According to the measurement of unknown classes, the metric evaluation can further be divided into distance modification and density distribution. 

Distance modification evaluates the distance margin between known and unknown classes. PEELER\cite{PEELER} integrates a meta-learning framework with Mahalanobis distance and introduces an open-set loss through negative entropy loss, aimed at maximizing the entropy of known classes. FSOSTC\cite{liFewShotOpenSetTraffic2022}
employs self-supervised learning to assist pre-training for the few-shot open-set traffic classification. openFEAT\cite{kishanOpenFEATImprovingSpeaker2022} collaboration with FEAT\cite{FEAT}, utilizing a set-to-set function, adapts speaker representations from a predefined universal embedding space to a family-specific embedding space, significantly enhancing the performance of household member identification. CTR\cite{dionelisCTRContrastiveTraining2022} proposes a unified framework for few-shot open-world recognition utilizing contrastive learning and employs negative training through distribution boundary contrastive loss. MRM\cite{cheBoostingFewShotOpenSet2023} proposes learning a distributional hypersphere for each class while simultaneously increasing the margin between different categories, implicitly delineating the distribution space of unknown classes.  ASOP\cite{kimTaskAgnosticOpenSetPrototype2023} proposes a task-agnostic open-set class with distance scaling factors, which is the nearest class for every sample except its corresponding ground-truth class. MSCL \cite{zhao2024meta} combines supervised contrastive learning and meta-learning principles to boost the open-set automatic modulation classification. G-FOOD \cite{su2024toward} proposes two innovative modules: the Class Weight Sparsification Classifier (CWSC) and the Unknown Decoupling Learner (UDL). CWSC reduces the co-adaptability between classes by randomly sparsifying the normalized weights for class logit prediction, while UDL decouples the training of the unknown class, enabling the model to form a compact unknown decision boundary.  

Density distribution is based on the assumption that unknown samples are located in low-density regions of the distribution, while known samples are in dense neighborhoods.
EVML\cite{palExtremeValueMetaLearning2023a} fits Weibull distributions for each known class based on distances from the limited support set to their respective prototypes. OSLO \cite{boudiafOpenSetLikelihoodMaximization2023a} applies a refined Maximum Likelihood principle and recalibrates likelihood values through a strategic weighting of samples, ensuring that samples categorized to known classes are assigned higher likelihoods. GEL\cite{wangGlocalEnergybasedLearning2023} leverages features
at both the class and pixel levels to learn energy scores. Specifically, global energy scores are computed using class-level features, capturing overall distinctions among samples and their predicted classes, while local energy scores concentrate on pixel-level details, uncovering fine-grained differences in features. 

\noindent\textbf{Summary and Discussion.} Metric evaluation methods optimize the feature space and enlarge the differences between known classes and unknown ones, enabling the distinction of unknown class samples even with a small support set. However, these methods struggle to establish clear distribution boundaries when the feature distributions of different classes are similar or overlapping. This ambiguity leads to confusion in distinguishing samples and ultimately reduces the model's recognition capability.

\vspace{-3mm}
\subsection{Feature Processing.} 
Feature processing, as depicted in Fig. \ref{fig:opensetExample}(d), reconstructs the feature and compares the differences between the reconstructed sets before and after to identify whether the sample belongs to an unknown class.
SnaTCHer \cite{SnaTCHer} proposes that the features of a sample will exhibit a high degree of consistency after undergoing specific transformation functions. SnaTCHer replaces the feature vector of each query sample with the predicted class prototype. The modified class prototype set is then reconstructed, and the sample is judged to match the predicted class prototype by comparing the differences of the set before and after reconstruction. Building on this, ATT\cite{huangTaskAdaptiveNegativeEnvision2022} proposes using transformation functions and other reconstruction techniques to generate prototypes representing unknown classes, which aims to provide a flexible rejection boundary for various few-shot open-set recognition tasks. Specifically, during the meta-testing, if a test query sample's predicted scores for all known classes are lower than its corresponding negative class prototype score, the classifier will reject the sample, deeming it to belong to an unknown class. ReFOCS \cite{nagReconstructionGuidedMetaLearning2023a} introduces reconstruction as an auxiliary task into the few-shot learning framework, guiding the classifier to develop a self-awareness ability. This unique self-awareness allows the model to naturally detect samples that likely belong to an unknown class when it struggles to accurately reconstruct the input. FreqDiMFT\cite{FreqDiMFT} introduces a module utilizing Fast Fourier Transform (FFT) to create a frequency distribution map from the original image and feature maps. This module captures domain-specific features by integrating frequency distribution information, addressing the high inter-class similarity and intra-class diversity in remote sensing images. RFDNet\cite{dengLearningRelativeFeature2023} learns to generate feature displacements for both the support set and the query adaptively based on information about the whole task, which is generated by transferring the knowledge extracted from the auxiliary dataset, i.e., linearly combining the base-class centroids followed by attentive modulation. 

\noindent\textbf{Summary and Discussion.}
Feature processing methods introduce additional information through feature reconstruction, establishing reference points with a limited number of known support samples, which can lead to more accurate measurements between known and unknown classes. 
However, these methods are hard to select the appropriate transformation function to ensure that the reconstructed features effectively represent the known class.

%% file: chapters/Incremental.tex
\section{Few-Shot Learning with Varying Classes in the Open World: Incremental Few-Shot Learning}
\begin{figure}[t]
\centering
    \includegraphics[width=0.95\columnwidth]{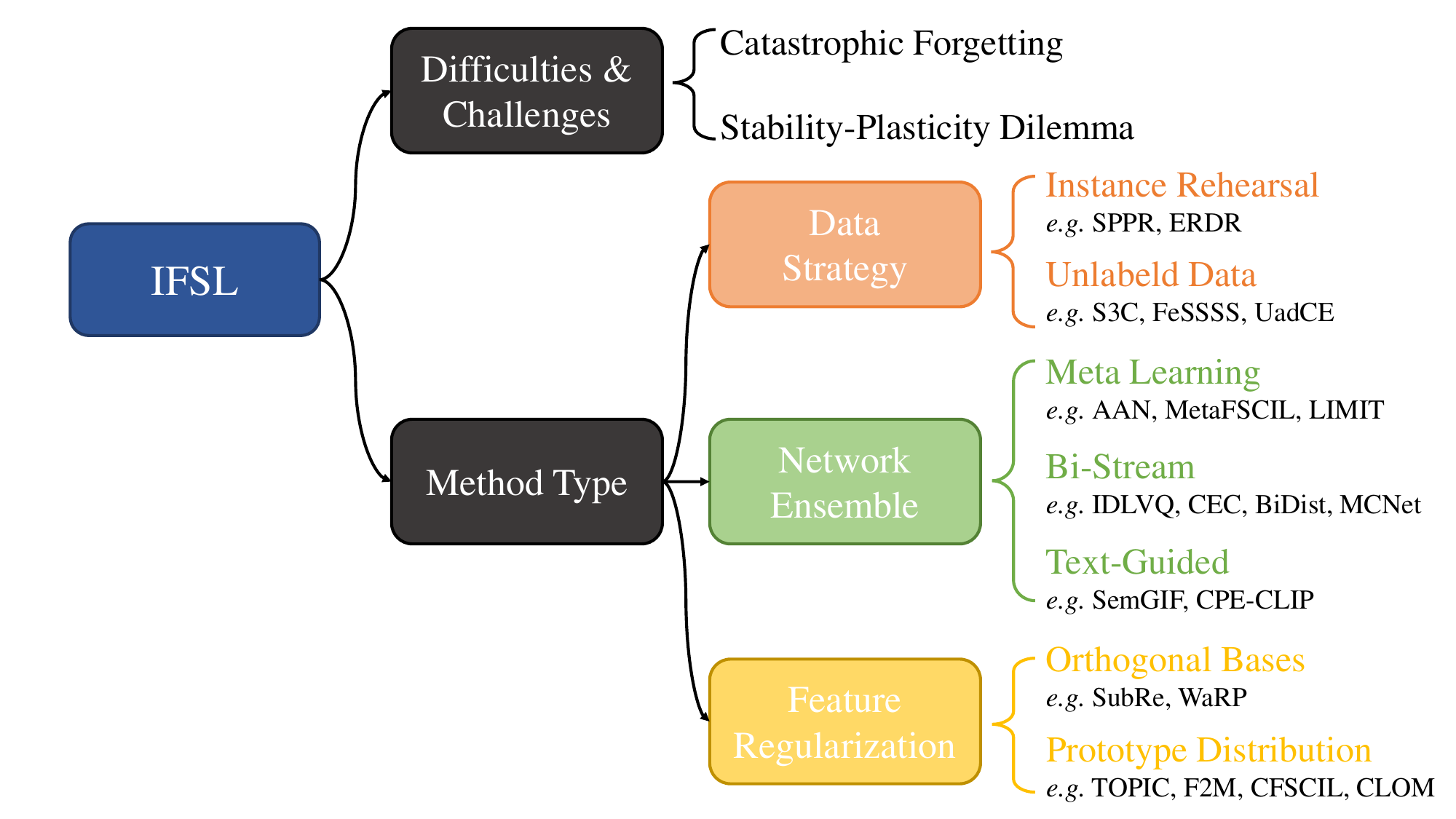}
    \caption{The structural relationships of the existing three types of methods and their subclasses, as well as the difficulties and challenges faced by IFSL problems.}
     \label{fig1:env}
      \vspace{-1mm}
\end{figure}

In the open-world environment, few-shot learning systems often face the challenge that classes of tasks as inputs vary frequently. The problem of few-shot learning with varying classes, also known as incremental few-shot learning, requires continuing to learn new tasks from novel input while retaining previously gained
knowledge. The key challenges of incremental few-shot learning come from two aspects: 1) catastrophic forgetting which refers to a significant decrease in the ability to complete old tasks while learning new ones.  Due to the high coupling of deep learning parameters, which makes it difficult to add new knowledge to neural networks alone without bringing about other changes, and 2) the stability-plasticity dilemma which is also worth considering. While fixed weights can retain old knowledge, they would lead to ineffective learning of new tasks, resulting in a loss of learning ability. To address the problem, existing works can be categorized into a unified taxonomy including Data Strategy, Network Ensemble, and Feature Regularization as shown in Fig. \ref{fig1:env}.  The data-based approach provides the system with the ability to backtrack knowledge through data manipulation. Common approaches include instance rehearsal (SPPR \cite{SPPR} and ERDR \cite{ERDR}) and the introduction of unlabeled data for self-supervised learning and semi-supervised learning (\cite{s3c, FeSSSS, UadCE}). Network-based methods aim to aggregate models with different performances. The mainstream technological paths include meta-learning (AAN \cite{AAN}, MetaFSCIL \cite{MetaFSCIL} and LIMIT \cite{LIMIT}), bi-stream networks (IDLVQ \cite{IDLVQ}, CEC \cite{CEC}, BiDist \cite{BiDist}, MCNet \cite{MCNet}), M2SD \cite{M2SD} and BH-RCNN \cite{BHRCNN}, and text guidance (SemGIF \cite{SemGIF} and CPE-CLIP \cite{CPECLIP}). Feature-based methods alleviate catastrophic forgetting by measuring the distribution in the feature space. They commonly including orthogonal basis (SubRe \cite{SubRe}, WaRP \cite{WaRP}) and OrCo \cite{OrCo} and prototype distribution(TOPIC \cite{TOPIC}, F2M \cite{F2M}, CFSCIL \cite{CFSCIL}, CLOM \cite{CLOM}, FACT \cite{FACT}, DSN \cite{DSN}, EHS \cite{EHS} and NCFSCIL \cite{NCFSCIL}). 

\begin{figure*}[t]
\centering
    \includegraphics[width=1.7\columnwidth]{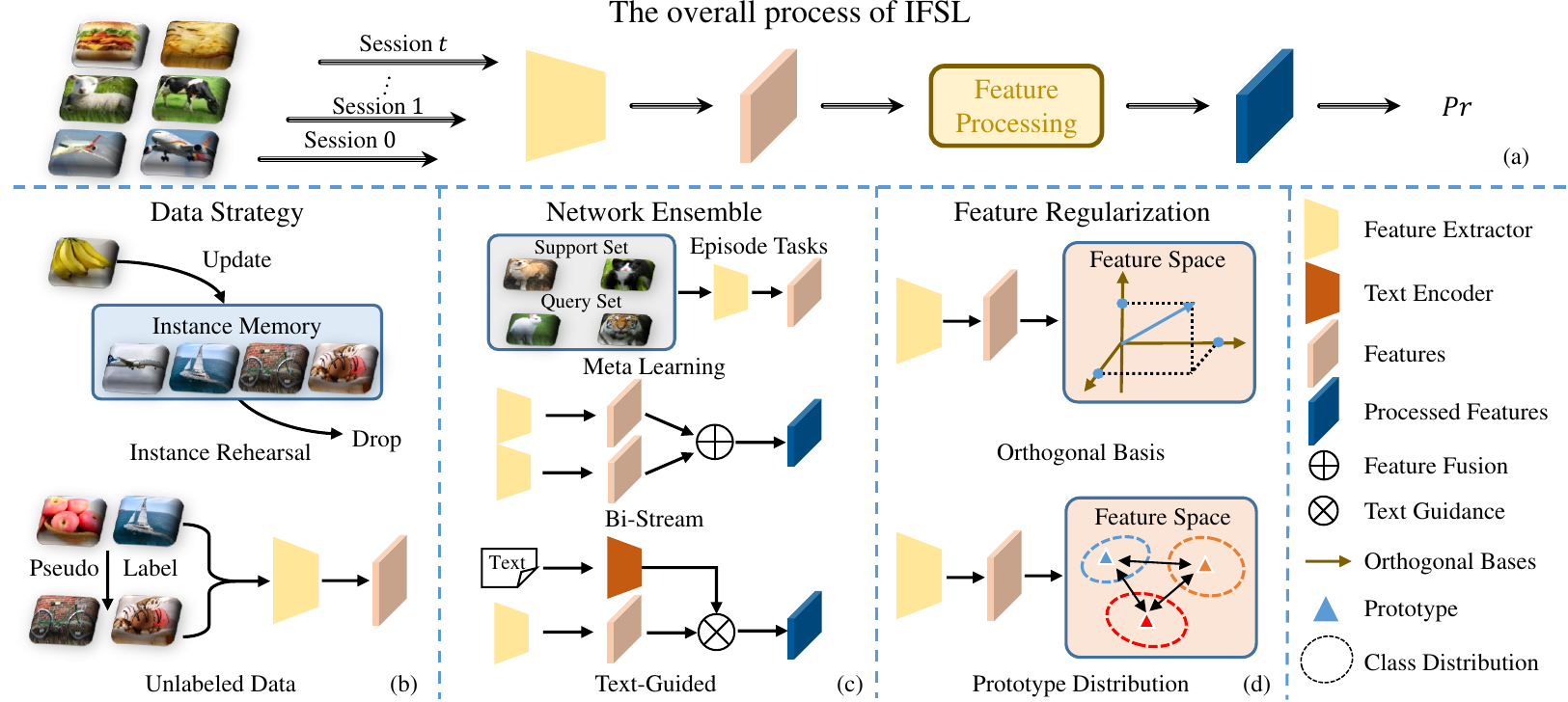}
    \caption{An overall diagram of existing IFSL researches. (a) Overview of the IFSL setting. (b) Data Strategy methods are mainly divided into Instance rehearsal and Unlabeled data. (c)Network Ensemble methods consider combining networks with different functions including Meta Learning networks that handle different tasks, Bi-Stream networks and Text-Guidance. (d) Feature Regularization methods perform incremental learning by constraining the distribution of instances in the feature space, commonly including Orthogonal Basis and Prototype Distribution.}
   \label{fig2:env}
    \vspace{-1mm}
\end{figure*}

In this section, we will introduce the latest progress of few-shot learning with varying classes from three aspects. The overview of the problem definition and method strategy is depicted in Fig. 
 \ref{fig2:env}.


\vspace{-3mm}
\subsection{Data Strategy} 
Data strategy, as depicted in Fig. \ref{fig2:env}(b), focuses on how to efficiently utilize limited labeled data, maintaining a dynamic memory set, and introducing unlabeled data are mainstream methods. According to the data usage method, data strategy can further be divided into instance rehearsal and unlabeled data. 

Instance rehearsal preserves representative samples in incremental learning for knowledge playback by maintaining a memory set. 
SPPR \cite{SPPR} proposes a random episode selection strategy for sampling and obtaining pseudo-incremental data to enhance the adaptability of the model. In addition, a dynamic relationship projection module is proposed based on the similarity of features before and after the model to ensure the stationarity of the feature space before and after updates. 
Unlike most previous studies, ERDR \cite{ERDR} believes that data replay can also be applied to IFSL tasks. Considering the issue of data privacy, the author proposes a rehearsal-free method that introduces entropy regularization to generate several pseudo samples near the decision boundary. This uncertainty can alleviate overfitting and forgetting problems in incremental learning.
Agarwal \textit{et al.} believe that it is not feasible to directly sample the memory set from the training set in IFSL, because the obtained examples cannot reflect the sample characteristics well due to data scarcity. They propose a method of instance rehearsal based on adversarial generation networks (GAN) to generate representative pseudo-samples.

Unlabeled data introduces unlabeled data in IFSL through unsupervised and semi-supervised learning, thus alleviating the problem of data scarcity. 
S3C \cite{s3c} proposes that the scarcity of data caused by few-shot is an important reason for forgetting, which can be mitigated by introducing self-supervised learning with unlabeled data. By randomly enhancing unlabeled images and proposing a random classifier to fuse the results of multiple classifiers, self-supervised learning is effectively applied to IFSL.
FeSSSS \cite{FeSSSS} proposes that self-supervised learning models can be used to assist IFSL. It believes that the main reason for the few-shot problem is the high labeling cost. Therefore, from a practical perspective, self-supervised models can be used to improve model performance. At the same time, the self-supervised data used does not need to be strictly distributed with the incremental learning data, and only needs to be roughly kept from the same domain.
UaD-CE \cite{UadCE} explores the application of semi-supervised learning in IFSL tasks, and its designed UaD module alleviates the interference of few-shot new tasks on pseudo labels by retaining the pseudo label information of the previous round, improving the quality of pseudo labels in semi-supervised learning. At the same time, UaD-CE designed the ClE module, which ensures the diversity of pseudo labels in the incremental learning process by applying class-balanced self-training (CB-ST).

\noindent \textbf{Summary and Discussion.}
Data strategy can mitigate catastrophic forgetting by using real samples for knowledge retention. However, there are some limitations to this approach. First, for the selection of the memory set, improper sampling methods will lead to the low quality of samples in memory, thus misleading the learning of the model. Second, in privacy information or edge computing scenarios, keeping a memory is considered undesirable, which limits the versatility of such methods.

\subsection{Network Ensemble}
Network ensemble, as depicted in Fig. \ref{fig2:env}(c), aims to combine different networks and parameters to adapt to constantly increasing new tasks. According to the difference of functions, network ensemble can further be divided into meta learning, bi-stream and text-guide methods. 

Meta learning methods employ episode training to integrate the ability to handle different tasks on a network. Unlike general ensemble learning, this is the integration at the level of network representation.
AAN \cite{AAN} first proposes the concept of few-shot incremental learning and solves this problem through meta-learning. AAN consists of three stages which are pre-training, attention attractor training, and meta incremental learning separately, and it has been proven that current backpropagation can effectively update the parameters of incremental models. 
MetaIFSL \cite{MetaFSCIL} introduces meta-learning into IFSL, it is proposed that task sequences can be obtained by sampling in the base class to simulate the incremental learning process. At the same time, a bidirectional guidance mechanism (BGM) is proposed, which can automatically calculate modulation maps on the support set to assist in query set classification. By sampling in the base class for pseudo incremental learning, LIMIT \cite{LIMIT} is prepared in advance for tasks that have not yet appeared. In addition, a transformer-based meta calibration method is proposed to align new prototypes and old classifiers before and after incremental learning.

Bi-Stream divides the network into two parallel branches, obtains the classification results from two perspectives respectively, and achieves the final output results through decision fusion. IDLVQ \cite{IDLVQ} only updates the correct prototype and the most recent incorrect prototype, alleviating the risk of overfitting caused by the few-shot setting. The proposed Learning Vector Quantization (LVQ) can be seen as a dynamic network architecture, as it continuously adds new reference vectors as incremental learning progresses. CEC \cite{CEC} proposes a three-stage network that considers new features as a combination of old features and calculates weighted weights through self-attention. It saves the features learned in each round as a memory for calculating Graph Attention Network (GAT), which means that the features have a contextual relationship and can consider the relationships between different features as the training rounds increase, then it is possible to learn this weighting method through meta-learning. 
BiDist \cite{BiDist} proposes that both the base model and the previous model can be used as teachers for bidirectional distillation, which can effectively alleviate the risk of overfitting during current task learning. At the same time, a bilateral network was introduced to predict the base class and new class, respectively, and the branch prediction results were merged through the attention aggregation module.
MCNet \cite{MCNet} obtains an ensemble learning model suitable for IFSL by decision fusion of feature extractors with different performances. When learning base classes, CNN and Transformer are used to train two independent feature extractors with different preferences for feature extraction, and attention regularization is used to fuse them. When learning new classes, a Prototype Smoothing Hard Mining Triplet (PSHT) loss is proposed, which achieves distinguishable feature distributions by sampling the distribution of new classes and constraining their boundaries.
M2SD \cite{M2SD} uses a dual branch structure during the base class training session, which promotes the base class to learn more robust feature representations through Multiple Mixing Self-Distillation mechanism, which is beneficial for subsequent incremental learning. This method enhances the ability of incremental learning by improving the generalization of the base class model, but it has special requirements for pre-training and cannot perform incremental learning on any existing model, which undoubtedly reduces the model's versatility.
BH-RCNN \cite{BHRCNN} separates the models of the base class and the novel class, and applies a method similar to ensemble learning to obtain a unified prediction result. Based on the proposed Recall and Progress mechanism, BH-RCNN achieves smooth movement of different category features in the incremental learning process, which is beneficial for alleviating catastrophic forgetting caused by overly scattered features.

Text-guide methods introduce text information to guide visual learning. SemGIF \cite{SemGIF} proposes that generating instances can replace instance memory. Conditional-GAN is applied in the instance generation process, which generates base class pseudo instances by sampling the support set to achieve data rehearsal. In addition, SemGIF uses label text embedding as guiding information for image semantics, thereby assisting in feature extraction. Furthermore, Alessandro \textit{et al.}\cite{CPECLIP} design the Continuous Parameter Efficient CLIP (CPE-CLIP) for text-vision multimodal IFSL based on CLIP. Unlike other methods, CLIP obtains a large number of strong feature representations on large-scale pre-training, so it does not require or only needs to update very few parameters in few-shot learning. By querying in the knowledge base, good feature representations can be obtained. CPE-CLIP significantly improves the performance of IFSL on multiple datasets, but comparing it with other methods that do not use pre-trained models is biased and poses a risk of data leakage.

\noindent \textbf{Summary and Discussion.}
Network ensemble methods have good scalability with multimodal information and can support both CIL and TIL settings. However, these methods are debatable whether the ensemble of networks can lead to the integration of knowledge. Training on few-shot tasks does not achieve comprehensive and sufficient training, which may result in insufficient knowledge integration and poor model generalization. Besides, due to the introduction of multi-model aggregation, such methods face fairness problems in comparison.

\vspace{-3mm}
\subsection{Feature Regularization} 
Feature regularization, as depicted in Fig. \ref{fig2:env}(d), aims to avoid mutual influence between incremental tasks by applying appropriate regularization. According to different types of regularization, this approach can be further divided into prototype distribution and orthogonal basis.

Prototype distribution methods measure the distance between different prototypes. TOPIC\textit{et al.}\cite{TOPIC} is presented to learn the topological structure of prototype vectors. TOPIC uses anchor loss when learning new tasks and minimizes the movement of old categories, which limits the distance of node movement during each update.  TOPIC also takes into account the topological relationship between prototypes. When adding a new category, it takes into account the distance between the new category and the existing categories, as well as the topological relationship between the nearest old class prototype and the new category.
F2M \cite{F2M} finds that many current IFSL algorithms do not perform as well as the base class model directly performing near class mean classification, due to the catastrophic forgetting problem in few-shot scenery, resulting in performance gain far worse than the performance drop. So it proposes to find a flat area in the base class model that is not sensitive to parameters, and learn new tasks in this area to slow down forgetting.
CLOM \cite{CLOM} finds that class level overfitting is the main cause of catastrophic forgetting, and models often overfit to base classes rather than specific samples. By introducing edges, an integrated model containing different edges was constructed, effectively improving the performance of IFSL.
C-IFSL \cite{CFSCIL} proposes another approach to explain and alleviate forgetting problems. It proposes that prototypes corresponding to different classes can be set to be pairwise orthogonal, in order to maintain the stability of the old task feature space when learning new tasks.
DSN \cite{DSN} proposes that the catastrophic forgetting problem of neural networks arises from highly coupled parameters, which further leads to the collapse of feature distributions for different categories. The DSN mechanism first expands the feature space to learn new knowledge, and then compresses parameters to retain old knowledge, effectively avoiding class confusion.
Yang \textit{et al.}\cite{NCFSCIL} draws inspiration from the phenomenon of neural collapse and proposes to introduce the optimal classification geometry structure in IFSL, which is a simplex equiangular tight frame (ETF) composed of class prototypes. By using dot-regression loss, the features are optimized towards a predetermined fixed class prototype direction to maintain the optimality of neural collapse during the incremental learning process.
FACT \cite{FACT} proposes that during training, preparations can be made in advance for future category expansion. It is assumed that there are several virtual classes, and when training the base classes, the target is specified as both real and virtual classes. In incremental learning, appropriate virtual classes are assigned to each newly added class to achieve consistency in the feature space before and after incremental learning. This method implies a hypothesis that the features required for base and novel categories are similar, so it is possible to prepare for the new class in advance when training the base class. Similar to CEC \cite{CEC}, FACT also has certain requirements for the training method of the base class, while TOPIC \cite{TOPIC} does not make any requirements and can be applied to any backbone network, which brings about differences in the universality of different methods.
By analyzing models trained based on CE loss, Song \textit{et al.}\cite{SAVC} finds that such models often have poor inter-class separability. Therefore, it is proposed to improve generalization performance by introducing supervised contrastive learning (SCL). Based on the MoCo \cite{MoCo} framework, Song \textit{et al.}\cite{SAVC} proposes the SAVC model, which introduces virtual categories during contrastive learning and reserves feature space positions in advance for possible new categories, thereby improving generalization, which is similar to FACT \cite{FACT}.
Similarly, EHS \cite{EHS} proposed an Extending Hypersphere Space method that utilizes odd-symmetric activation functions to avoid incompleteness in the feature space, while reserving feature space for future new classes. 
Wang \textit{et al.}\cite{TEEN} propose a Training frEE scalability calibratioN (TEEN) strategy by using base class prototypes as anchors to adjust new class prototypes.
MICS \cite{MICS} has designed an IFSL method that is consistent between base and novel classes, and through experiments, it has been found that the Mixup \cite{Mixup} method has a promoting effect on alleviating the semantic gap between novel and base classes and enhancing intra-class compactness and inter-class separability in metric learning.

Orthogonal basis methods decompose the feature vector into a set of orthogonal basis vectors, thereby avoiding interfering with old knowledge when learning new tasks. SubRe \cite{SubRe} proposes to perform orthogonal decomposition on the base class features to obtain a set of bases, and use the convex combination of bases as the new class feature representation, thereby ensuring that the model learns semantic features instead of spurious features. In addition, guided by label semantics, a high-confidence optimization direction was selected for the model.
Similar to \cite{CFSCIL}, WaRP \cite{WaRP} finds that the catastrophic forgetting in incremental learning is due to the high coupling of neural network parameters, and even fine-tuning very few parameters can change most of the previous knowledge. Therefore, the weight space rotation method WaRP is proposed, which improves the forgetting of previous knowledge by finding a set of orthogonal bases in the weight space and only fine-tuning them in their smoothest direction. 
OrCo \cite{OrCo} also utilized a method similar to FACT \cite{FACT} by preparing virtual class prototypes for future categories in advance, with the difference being that it designed a series of mutually orthogonal prototypes as pseudo targets. In addition to cross-entropy loss, OrCo also aligns instance features and pseudo targets by applying orthogonal loss and perturbed supervised contrastive loss, improving the separability of instances in the feature space.

\noindent \textbf{Summary and Discussion.}
The methods focus on the distribution of feature space and alleviate catastrophic forgetting through regularization. They take into account the global relationship of feature distributions, making the feature representation consistent among different tasks. However, it is difficult to disentangle the effects of large numbers of coupling parameters during regularization, which makes knowledge retention very difficult. Besides, feature regularization requires clear distribution boundaries between categories, but this is often difficult to achieve under few-shot conditions. 

%% file: chapters/cross-domain/Cross-domain.tex
\section{Few-Shot Leaning with Varying Distribution in the Open World: Cross-Domain Few-Shot Learning}

\begin{figure}
\centering
    \includegraphics[width=0.95\columnwidth]{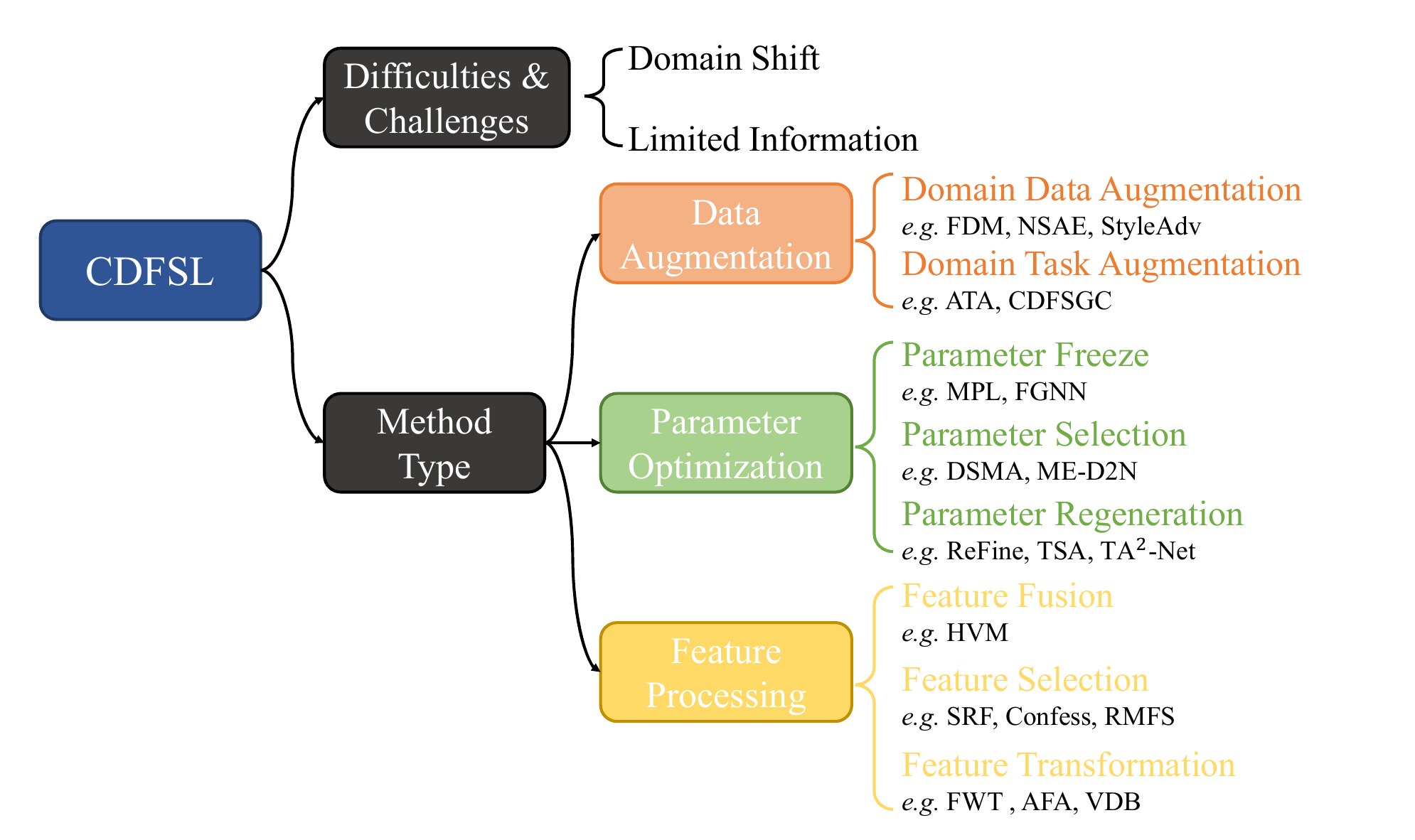}
    \caption{The structural relationships of the existing three types of methods and their subclasses, as well as the difficulties and challenges faced by CDFSL problems.}
    \label{fig_CD_1:tree}
     \vspace{-2mm}
\end{figure}

In the open-world environment, few-shot learning systems often face the challenge that distributions from source domains to target domains vary frequently. The problem of few-shot learning with varying distribution, also known as cross-domain few-shot learning, requires to transfer the knowledge learned on the source domain to the target domain. 
The key challenges of Cross-Domain few-shot learning come from two aspects: 1) domain shift: there are considerable gaps between source and target domains, and 2) limited information: the scarcity of sample sizes in the target domain makes the cross-domain problems more challenging.
Therefore, models trained under the CDFSL problem need to have robust and strong generalization capabilities for source domains adapting to novel target domains with limited examples. To address this issue, several works have focused on adapting to the cross-domain environment, which can be categorized into a unified taxonomy including Data Augmentation, Parameter Optimization and Feature Processing, as shown in Fig \ref{fig_CD_1:tree}. The data-based approach focuses on augmenting data as well as tasks. Common approaches include data enhancement (e.g. FDM \cite{c3_fu2021meta}, NSAE \cite{c39_liang2021boosting} and StyleAdv \cite{c9_fu2023styleadv}) and task enhancement (e.g. ATA \cite{c4_wang2021cross} and CDFSGC \cite{c10_hassani2022cross}). The parameter-based approach is widely used in CDFSL. The mainstream methods include parameter freeze (e.g. MPL \cite{c33_wang2021meta}, and FGNN \cite{c18_chen2022cross}), parameter selection (e.g. DSMA \cite{c79_tu2021dropout} and ME-D2N \cite{c41_fu2022me}), and parameter regeneration (e.g. ReFine \cite{c30_oh2022refine}, TSA \cite{c31_li2022cross} and TA$^2$-Net \cite{c44_guo2023task}). Feature-based methods perform different operations on features such as feature fusion (e.g. HVM \cite{c13_du2021hierarchical}), feature selection (e.g. SRF \cite{c15_zou2021revisiting}, Confess \cite{c16_das2021confess} and RMFS \cite{c17_weng2021representative}), and feature transformation (e.g. FWT \cite{c28_tseng2020cross}, AFA \cite{c5_hu2022adversarial} and VDB \cite{c27_yazdanpanah2022visual}). 

\begin{figure*}
\centering
    \includegraphics[width=1.7\columnwidth]{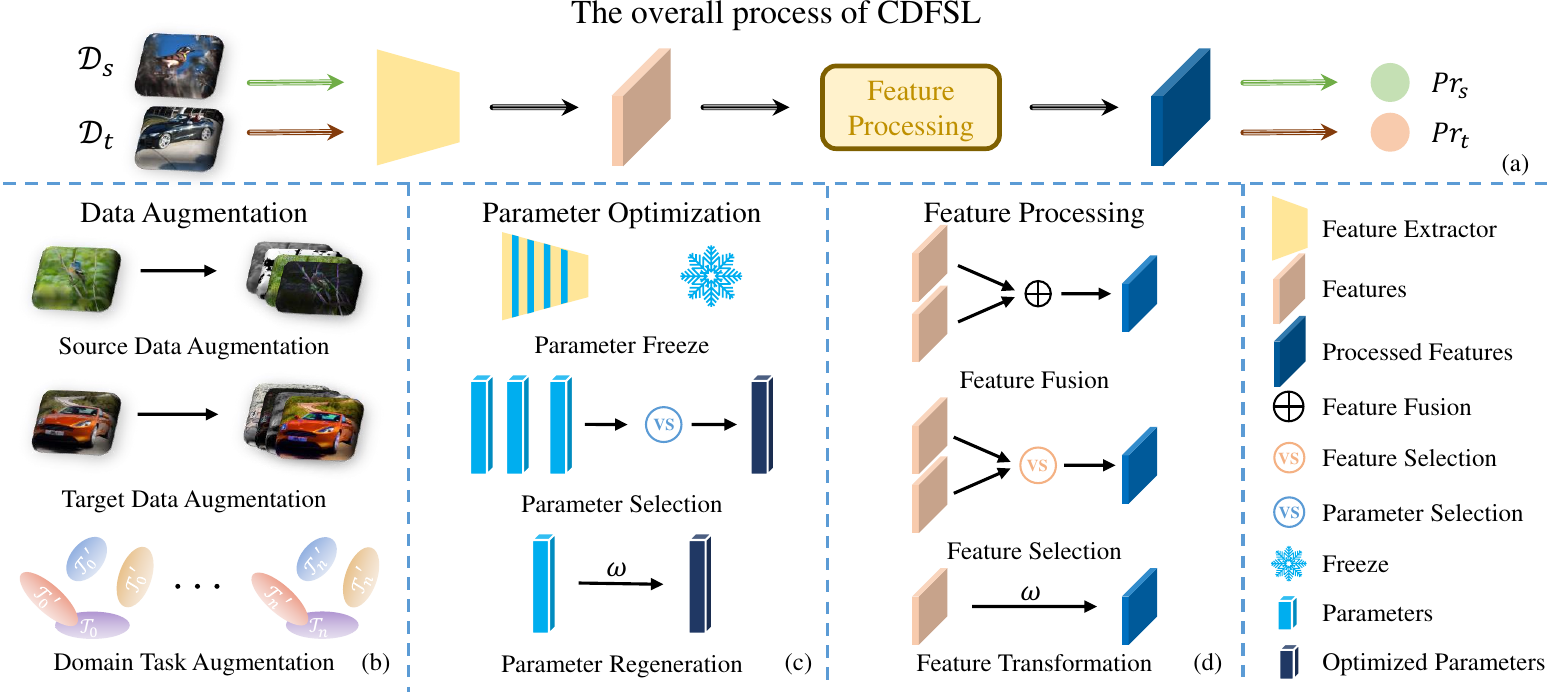}
    \caption{An overall diagram of existing CDFSL researches. (a) Overview of the CDFSL setting. (b) Data Augmentation methods are mainly divided into  Domain Data Augmentation(Source and Target) and Domain Task Augmentation. (c) Parameter Optimization Methods contain Parameter Freeze, Parameter selection and Parameter Regeneration. (d)Feature Processing methods consider Feature Fusion, Feature Selection and Feature Transformation. }
    \label{figCD:env}
     \vspace{-2mm}
\end{figure*}

In this section, we will introduce the latest progress of few-shot learning with unknown distributions from three aspects. The overview of the problem definition and method strategy is depicted in Fig. \ref{figCD:env}.

\vspace{-3mm}
\subsection{Data Augmentation}
Data Augmentation, as depicted in Figure \ref{figCD:env} (b), enhances the size and quality of training datasets for deep learning models especially facing Few-Shot Learning (FSL) problems \cite{c2_shorten2019survey}. According to different types of data, data augmentation can further be divided into domain data augmentation and domain task data augmentation.

Domain data augmentation aims to boost the training samples to extract more general knowledge to improve the generalization performance of the model over the target domain. The augmentation data includes source and target data. Feature Distribution Matching (FDM) \cite{c3_fu2021meta} proposes a feature-wise domain adaptation module to augment the query set samples by the MixUp process, which encourages the model to generate more synthetic samples for better domain adaptation. NSAE \cite{c39_liang2021boosting} trains the model by jointly reconstructing inputs and predicting the labels of inputs as well as their reconstructed pairs. StyleAdv \cite{c9_fu2023styleadv} is proposed to augment source domain samples by a novel style adversarial attack method. TACDFSL \cite{c7_zhang2022tacdfsl} and Deng \textit{et al.} \cite{c8_deng2022improving} apply rotation transformations to augment images and predict the rotation angle in the pretrain phase. Similarly, CDTA \cite{C22_li2023cdta} corrupts the semantic information of the benign image by scrambling the outputs of both the intermediate feature layers and the final layer of the feature extractor.

Domain task augmentation takes the task as a whole and tries to generate more tasks. For example, ATA \cite{c4_wang2021cross} considers the worst-case problem around source domain tasks and proposes adversarial methods to augment the source domain task distribution to address the domain gap in few-shot learning. In CDFSGC \cite{c10_hassani2022cross}, by augmenting the graphs from sampled tasks into three views, the representation of graphs is obtained for fast adaptation as well as knowledge transfer. Furthermore, by comparing predictions of weakly-augmented unlabeled target data from a teacher network to strongly-augmented versions of the same images from a student network \cite{c11_islam2021dynamic}, consistency regularization is imposed to learn a representation that can be easily adapted to the target domain. 

\noindent \textbf{Summary and Discussion.}
Data augmentation is a simple but effective technique that can be flexibly used in CDFSL to expand the amount of data. However,  the improvement of their model performance is not significant since the fundamental issue of domain shift has not been changed. The domain gap between the source and target domains is still significant. Therefore, compared to the other two methods that are mentioned below, data augmentation can probably be seen more as a trick than as a core.

\vspace{-3mm}
\subsection{Parameter Optimization}
Parameter optimization, as depicted in Figure \ref{figCD:env} (c), improves the model performance by fine-tuning model parameters. According to different parameter handling methods, parameter optimization can further be divided into parameter freeze, parameter selection and parameter regeneration. 

Parameter freeze methods fix some model parameters to limit the complexity of the hypothesis space. Among them, MPL \cite{c33_wang2021meta} proposes meta-prototypical learning to alternately freeze the meta-encoder's parameters to learn the general features. Moreover, FGNN \cite{c18_chen2022cross} proposes two new strategies, respectively for the encoder and the metric function of metric-based network.

Parameter selection strategy tries to identify the most suitable set of parameters for better adaptation to the target domains. For example, DSMA \cite{c79_tu2021dropout} samples several sub-networks by dropping neurons (or feature maps) to construct a bunch of models with diverse features for the target domain and chooses the most suitable sub-networks to construct the ensemble for the target domain learning. ME-D2N \cite{c41_fu2022me} introduces few labeled target domain data and transfers the knowledge from two teachers to a unified student model by the knowledge distillation technique.

The parameter regeneration technique adjusts some model parameters to further optimize the performance of the model. Various studies have been explored to boost cross-domain ability. For instance, ReFine \cite{c30_oh2022refine} re-randomizes the parameters fitted on the source domain before adapting to the target data, which resets source-specific parameters of the source pre-trained model and thus facilitates fine-tuning on the target domain. TSA \cite{c31_li2022cross} proposes to learn task-specific weights from scratch directly on a small support set, in contrast to dynamically estimating them. ProLAD \cite{c40_yang2024leveraging} utilizes two separate adapters with one devoid of a normalization layer, which is more effective for similar domains, and another embedded with a normalization layer, designed to leverage the batch statistics of the target domain, thus proving effective for dissimilar domains. TA$^2$-Net \cite{c44_guo2023task} is trained by reinforcement learning to adaptively estimate the optimal task-specific parameter configuration for each test task. UCDFSL \cite{c46_oh2022understanding} investigates which pre-training is preferred based on domain similarity and few-shot difficulty of the target domain and designs two pre-training schemes, mixed-supervised and two-stage learning. MetaPrompt \cite{c55_wu2024task} jointly exploits prompt learning and the parameter generation framework to flexibly produce a task-adaptive prompt with arbitrary length for unseen tasks. SWP \cite{c56_ji2024soft} combines the soft weight pruning strategy and regularization to effectively restrict redundant weights while simultaneously learning crucial features for both source and target tasks. FLUTE \cite{c77_triantafillou2021learning} designs a separate network that produces an initialization of parameters for each given task, and we then finetune its proposed initialization via a few steps of gradient descent. 

\noindent \textbf{Summary and Discussion.}
Parameter Optimization is critical to improve the model's performance which can find the most suitable set of parameters. However, most methods utilize simple structures or have a limited selection of the parameters, resulting in limited improvement of the model's performance.

\vspace{-3mm}
\subsection{Feature Processing}
Feature processing aims at achieving the most transferable feature representation by processing the original features. According to different feature processing methods, feature processing can further be divided into feature fusion, feature selection and feature transformation.

Feature fusion focuses on fusing features from different depths of the neural network or from different domains or different dimensions within a single feature. HVM \cite{c13_du2021hierarchical} uses a hierarchical prototype model to flexibly rely on features at different semantic levels which enables the model to adaptively choose the most generalizable features. Likewise, Zou \textit{et al.} \cite{c15_zou2021revisiting} fuse mid-level features to learn discriminative information of each sample and thus boost the discriminability of the model. FDFSL \cite{c47_qin2024cross} employs an orthogonal low-rank feature disentanglement method to acquire desired features of source and target pipelines.

Feature selection strategy often selects domain-general features to aid target domain generation. For example, In order to obtain the more relevant domain feature, SRF \cite{c15_zou2021revisiting} uses a multi-domain feature bank to automatically select the most relevant representations. Confess\cite{c16_das2021confess} utilizes a contrastive learning and feature selection system to address domain gaps between base and novel categories. In addition, the RMFS \cite{c17_weng2021representative} algorithm together with multi-domain feature selection is proposed to optimize the multi-domain features. MMT \cite{c52_wang2023mmt} bridges the domain gap between source and target domains, leveraging style-memory and content-memory components. LRP \cite{c54_sun2021explanation} develops a model-agnostic explanation-guided training strategy that dynamically finds and emphasizes the features which are important for the predictions. 

At the same time, feature transformation reweights features when extra network and multi-domain data are not available. A Feature-Wise Transformation (FWT) layer \cite{c28_tseng2020cross} is added after the BN layer in convolutional neural networks to simulate feature distributions in different domains. Simple CNAPS \cite{c78_bateni2020improved} learns adaptive feature extractors that allow useful estimation of the high dimensional feature covariances from surprisingly few samples. RDC-FT \cite{c23_li2022ranking} constructs a non-linear subspace to minimize task-irrelevant features therewithin while keeping more transferrable and discriminative information by a hyperbolic tangent transformation. Wang \textit{et al.} \cite{c25_wang2022experiments} and Xu \textit{et al.} \cite{c36_xu2024enhancing} both normalize features to alleviate feature dissimilarity. VDB \cite{c27_yazdanpanah2022visual} transfers the intermediate feature using the source and target domain statistical elements. IM-DCL \cite{c36_xu2024enhancing} employs a weighted distance calculation among features to establish a soft classification of the positive and negative sets for the entire feature set. ProD \cite{c43_ma2023prod} uses the prompting mechanism in the transformer to disentangle the domain-general (DG) and domain-specific (DS) knowledge from the backbone feature. MFIE \cite{C49_lin2024diversifying} uses LMMs to automatically translate class-specific images into class-agnostic natural language descriptions for various key semantic attributes in target domains and edit origin images based on class-agnostic natural language descriptions. zxVAD \cite{C51_aich2023cross} uses a novel Normalcy Classifier module to learn the features of normal event videos by learning how such features are different `relatively' to features in pseudo-abnormal examples. PMNet \cite{c53_chen2024pixel} extracts the domain-agnostic pixel-level affinity matching with a frozen backbone and captures both the pixel-to-pixel and pixel-to-patch relations in each support-query pair with the bidirectional 3D convolutions. Another work \cite{c5_hu2022adversarial} introduces a plug-and-play adversarial feature augmentation (AFA) method for CDFSL. Zhang \textit{et al.} \cite{c19_zhang2022dual} propose to use feature- and distribution-level cross-domain graph alignments to mitigate the impact of domain shift on CDFSL. Chen \textit{et al.} \cite{c20_chen2022cross} learn prototypical compact and cross-domain aligned representations, so that the domain shift and few-shot learning can be addressed simultaneously. URL \cite{c21_li2021universal} learns a single set of universal visual representations by distilling knowledge of multiple domain-specific networks after co-aligning their features with the help of adapters and centered kernel alignment. MvSA \cite{c38_zhao2024deep} makes use of labeled source domain data and easy target domain data to perform target-to-source and target-to-target alignment. CDSTN \cite{c48_zhang2023cross} introduces a set of domain classifiers behind each residual block for the purpose of domain alignment by extracting more domain information at different depths of the network. Dara \cite{C50_zhao2023dual} proposes a prototypical feature alignment to recalibrate support instances as prototypes and reproject these prototypes with a differentiable closed-form solution. 
AFGR \cite{c34_sa2022attentive} enhances different semantic feature information to help the metric function better locate the fine-grained feature information of the image.

\noindent \textbf{Summary and Discussion.}
Feature processing strategy is beneficial to preserve or generate the most suitable features for the target domain. However,  feature processing may introduce additional computational costs due to additional parameters. Besides, feature processing may only allow limited exploration for most process features from the final layer.

%% file: chapters/Performances.tex
\section{Performances}
\subsection{Few-shot Learning with Noisy Instances}
\subsubsection{Datasets and Settings}
To ensure a fair comparison of the NFSL approaches, we select some representative NFSL methods and compare them with the same implementation requirements and configuration. We conduct two widely used few-shot image recognition benchmarks: \textit{mini}ImageNet \cite{MatchingNet} and \textit{tiered}ImageNet \cite{tImageNet}.
\begin{itemize}
\item \textit{mini}ImageNet \cite{MatchingNet} contains 64, 16, and 20 classes for training, validation, and testing set, with 60K images in total. 
\item \textit{tiered}ImageNet \cite{tImageNet} contains 351, 97, and 160 classes for training, validation, and testing sets, respectively, with a total of approximately 0.78 million images.
\end{itemize}

In this paper, we focus on exploring symmetric label swap noise, which means the noise comes from the current task, it is produced by replacing the raw sample with a sample from the classes of the episode. Following previous work \cite{TraNFL}, we add noise to each class uniformly, in other words, meaning that the same proportion of noise is added for all classes.
Parameter optimization methods are devoted to alleviating the noise in the meta-training datasets instead of the meta-testing datasets. Sample selection methods focus on training the meta-learner to adapt to the noisy support sets in the test stage.

\subsubsection{Metric}
In NFSL, the metric is the accuracy under different noise rates. Specifically, we explore three types of noise rates: 0\%, 20\%, 40\%. In the support set, the number of noise samples per class is equal to the noise ratio multiplied by the number of samples per class.

\subsubsection{Performance Comparison and Analysis}
Table \ref{NFSL_results} summarizes the performance of selected NFSL approaches
evaluated on the two benchmark datasets, \textit{mini}Imagenet and \textit{tiered}ImageNet. The results of the methods based on parameter optimization are obtained from the original papers, while the results of the methods based on sample selection are from our version of implementation.

Table \ref{NFSL_results} presents the performance of compared methods faced with various noise rates of symmetric noise on two datasets. From Table \ref{NFSL_results}, it can be observed that, for the parameter optimization-based methods, compared to PCL, Nested-MAML shows inferior performance because its model is based on gradient-based methods and PCL is the model based on metric learning. Gradient-based methods commonly require multiple gradient updates, resulting in high computational costs and a tendency to get stuck in local optima. In contrast, metric-based methods are more efficient in training and possess strong generalization capabilities, leading to superior performance. Therefore, PCL achieves better performance. Meanwhile, in the sample selection-based methods, among the meta-training weighting methods, IDEAL achieves the best performance. For the test-time adaptation methods, DETA achieves the best performance. Comparing DETA and IDEAL, we observe that DETA generally outperforms IDEAL in most cases, indicating that test-time adaptation methods with powerful backbones currently hold a certain advantage over meta-training methods.

\begin{table*}[htbp]
  \centering
  \caption{The NFSL performance of the proposed methods on \textit{mini}ImageNet and \textit{tiered}ImageNet benchmarks. $K$ means 5-way $K$-shot.}
	\resizebox{1.925\columnwidth}{!}{
    \begin{tabular}{ccccccccccc}
    \toprule
    \multicolumn{1}{c}{\multirow{2}[1]{*}{Type}} & \multirow{2}[1]{*}{Methods} & \multirow{2}[1]{*}{Venue} & \multirow{2}[1]{*}{Backbone} & \multirow{2}[1]{*}{K-shot} & \multicolumn{3}{c}{\textit{mini}ImageNet} & \multicolumn{3}{c}{\textit{tiered}ImageNet} \\
     \cmidrule{6-11}      &       &       &       &       & 0\%   & 20\%  & 40\%  & 0\%   & 20\%  & 40\% \\
    \midrule
    \multicolumn{1}{c}{\multirow{3}[0]{*}{Parameter Optimization }} & Nested-MAML\cite{NESTEDMAML} & AAAI22 & ConvNet4 & 5     & 61.2$\pm$0.21 & 59.6$\pm$0.54 & 56.46$\pm$0.74 &   -    &    -   &  -  \\

          & PCL\cite{PCL}   & IJCAI23 & ConvNet4 & 5     & 70.31$\pm$0.69 & 68.70$\pm$0.72 & 67.64$\pm$0.68 & 69.94$\pm$0.81 & 68.95$\pm$0.75 & 66.69$\pm$0.77 \\
        \midrule
    \multicolumn{1}{c}{\multirow{5}[1]{*}{Sample Selection}} & RNNP\cite{RNNP}  & WACV21 & ConvNet4 & 5     & 68.38$\pm$0.24 & 62.43$\pm$0.25 & 51.62$\pm$0.23 & 71.36$\pm$0.18 & 65.95$\pm$0.19 & 54.86$\pm$0.21 \\
          & RapNets\cite{RFSL} & TNNLS21 & ConvNet4 & 5     & 66.45$\pm$0.25 & 63.10$\pm$0.26 & 55.08$\pm$0.25 & 68.02$\pm$0.25 & 64.61$\pm$0.26 & 56.85$\pm$0.25 \\
          & TraNFS\cite{TraNFL} & CVPR22 & ConvNet4 & 5     & 68.13$\pm$0.23 & 64.45$\pm$0.25 & 56.76$\pm$0.25 & 69.63$\pm$0.23 & 65.76$\pm$0.25 & 56.96$\pm$0.25 \\
          & IDEAL\cite{IDEAL} & TPAMI23 & ConvNet4 & 5     & 67.93$\pm$0.22 & 64.27$\pm$0.24 & 56.45$\pm$0.22 & 71.07$\pm$0.22 & 66.97$\pm$0.22 & 60.13$\pm$0.23 \\
          & DETA\cite{DETA}  & ICCV23 & ConvNet4 & 5     & 70.21$\pm$0.22 & 65.93$\pm$0.21 & 57.04$\pm$0.22 & 72.41$\pm$0.23 & 68.77$\pm$0.22 & 59.57$\pm$0.22 \\
    \bottomrule
    \end{tabular}
    }%
  \label{NFSL_results}%
    \vspace{-3mm}
\end{table*}%

\vspace{-3mm}
\subsection{Few-shot Learning with Unknown Instances}
\subsubsection{Datasets and Settings} To compare the FSOSR approaches in a fair manner, We select some representative IFSL methods and compare them in the same implementation requirements
and configurations. FSOSR experiments are commonly conducted on two datasets, \textit{mini}ImageNet\cite{MatchingNet} and \textit{tiered}ImageNet\cite{tImageNet}.

In terms of open-set settings, following existing researchers\cite{PEELER}, we set $N$ = 5 and $K$ = 1, 5 during meta-training and meta-testing. For each episode, we sample 5 known classes and 5 unknown classes, with each containing 15 queries. All support set samples are from the known classes. The query set contains samples from known classes, the known query set, and samples from unknown classes, the unknown query set.

\subsubsection{Metrics} In FSOSR, two metrics are applied to measure the performance of the model, including closed-set classification accuracy (Acc) and Area Under the Receiver Operating Characteristics curve (AUROC) for unknown class detection. The Acc measures the classification capacity via known samples, and the AUROC evaluates the unknown detection capacity via both known and unknown samples.

\subsubsection{Performance Comparison and Analysis} The experimental results on the \textit{mini}ImageNet and \textit{tiered}ImageNet datasets are counted as shown in Table.\ref{tab:openset}. All the results come from the original papers. 

Among the three types of methods, metric-based approaches consistently exhibit superior stability and performance compared to the other two types. 
Specifically, GEL\cite{wangGlocalEnergybasedLearning2023} stands out as the top performer across all methods. GEL learns global open scores for detecting unknown samples by analyzing class-wise and pixel-wise scales. It assigns high energy scores to samples that significantly differ from the few-shot examples in either class-wise features or pixel-wise features, and assigns low energy scores otherwise. In the data-based method, proCAM\cite{songFewshotOpensetRecognition2022} and its variant exhibit remarkable performance by leveraging background regions of images from known classes, thus adding auxiliary information from known samples.  It reveals that generating appropriate virtual samples can simulate the open environment during testing, thereby indicating the ability of few-shot models to detect unknown class samples.
In the metric-based methods, apart from GEL, MRM\cite{cheBoostingFewShotOpenSet2023} and ASOP\cite{kimTaskAgnosticOpenSetPrototype2023} also demonstrate strong performance, which might be attributed to their effective optimization of the feature space.  The excellent performances of these approaches demonstrate the efficient generalization capability of metric-based methods. In the feature-based methods, ATT\cite{huangTaskAdaptiveNegativeEnvision2022} and its variant ATT-G showcase good performance, which generates negative prototypes to represent unknown classes through the transformation function, thus providing a flexible and concise rejection boundary. Meanwhile, ATT-G utilizes class semantics to model the unknown class and achieve better performance than other methods.

\begin{table*}[htbp]
  \centering
  \caption{The FSOSR performance of the proposed methods on \textit{mini}ImageNet and \textit{tiered}ImageNet benchmarks. 1-shot, 5-shot means 5-way 1-shot and 5-way 5-shot.}
	\resizebox{1.925\columnwidth}{!}{
    \begin{tabular}{cccccccccccc}
    \toprule
    \multicolumn{1}{c}{\multirow{3}[6]{*}{Type}} & \multirow{3}[6]{*}{Methods} & \multirow{3}[6]{*}{Venue} & \multirow{3}[6]{*}{Backbone} & \multicolumn{4}{c}{\textit{mini}ImageNet} & \multicolumn{4}{c}{\textit{tiered}ImageNet} \\
\cmidrule{5-12}          &       &       &       & \multicolumn{2}{c}{1-shot} & \multicolumn{2}{c}{5-shot} & \multicolumn{2}{c}{1-shot} & \multicolumn{2}{c}{5-shot} \\
\cmidrule{5-12}          &       &       &       & Acc   & AUROC & Acc   & AUROC & Acc   & AUROC & Acc   & AUROC \\
    \midrule
    \multicolumn{1}{c}{\multirow{2}[1]{*}{Data Augmentation}} & ProCAM\cite{songFewshotOpensetRecognition2022} & ACM MM22 & ResNet12 & 67.78 & 71.41 & 83.45 & 78.19 & 68.16 & 75.35 & 86.34 & 82.76 \\
          & ProCAMsm\cite{songFewshotOpensetRecognition2022}& ACM MM22 & ResNet12 & 67.86 & 71.09 & 83.66 & 77.51 & 68.82 & 75.55 & 85.64 & 82.77 \\
    \midrule
    \multicolumn{1}{c}{\multirow{6}[1]{*}{ Metric Evaluation}} & PEELER\cite{PEELER} & CVPR20 & ResNet12 & 65.86±0.85 & 60.57±0.83 & 80.61±0.59 & 67.35±0.80 & 69.51±0.92 & 65.20±0.76 & 84.10±0.66 & 73.27±0.71 \\
          &  MRMC-ProtoNet\cite{cheBoostingFewShotOpenSet2023} & IJCAI23 & ResNet12 & 64.05±0.82 & 68.11±0.73 & 84.73±0.51 & 77.07±0.63 & 68.35±0.94 & 72.99±0.74 & 84.75±0.62 & 78.18±0.60 \\
          &  MRMT-SnaTCHerF\cite{cheBoostingFewShotOpenSet2023} & IJCAI23 & ResNet12 & 67.03±0.83 & 71.20±0.80 & 82.00±0.55 & 80.39±0.59  & 71.13±0.91  & 75.59±0.77  & 85.27±0.62 & 83.03±0.63 \\
          & ASOP\cite{kimTaskAgnosticOpenSetPrototype2023}  & ICIP23 & ResNet12 & 66.30±0.60 & 71.72±0.50 & 82.68±0.20 & 80.92±0.20 & 68.68±0.70 & 74.92±0.40 & 84.84±0.40 & 82.37±0.30 \\
          & ASOP-L\cite{kimTaskAgnosticOpenSetPrototype2023} & ICIP23 & ResNet12 & 67.85±0.20 & 71.91±0.70 & 82.81±0.30 & 81.04±0.20 & 71.49±0.40 & 75.04±0.40 & 85.15±0.10 & 81.51±0.10 \\
          & GEL\cite{wangGlocalEnergybasedLearning2023}   & CVPR23 & ResNet12 & 68.26±0.85 & 73.70±0.82 & 83.05±0.55 & 82.29±0.60 & 70.50±0.93 & 75.86±0.81 & 84.60±0.65 & 81.95±0.72 \\
    \midrule
    \multicolumn{1}{c}{\multirow{6}[2]{*}{Feature Processing}} & SnaTCHe-F\cite{SnaTCHer} & CVPR21 & ResNet12 & 67.02±0.85 & 68.27±0.96 & 82.02±0.53 & 77.42±0.73 & 70.52±0.96 & 74.28±0.80 & 84.74±0.69 & 82.02±0.64 \\
          & SnaTCHe-T\cite{SnaTCHer} & CVPR21 & ResNet12 & 66.60±0.80 & 70.17±0.88 & 81.77±0.53 & 76.66±0.78 & 70.45±0.95 & 74.84±0.79 & 84.42±0.68 & 82.03±0.66 \\
          & SnaTCHe-L\cite{SnaTCHer} & CVPR21 & ResNet12 & 67.60±0.83 & 69.40±0.92 & 82.36±0.58 & 76.15±0.83 & 70.85±0.99 & 74.95±0.83 & 85.23±0.64 & 80.81±0.68 \\
          & ATT\cite{huangTaskAdaptiveNegativeEnvision2022}   & CVPR22 & ResNet12 & 67.64±0.81 & 71.35±0.68 & 82.31±0.49 & 79.85±0.58 & 69.34±0.95 & 72.74±0.78 & 83.82±0.63 & 78.66±0.65 \\
          & ATT-G\cite{huangTaskAdaptiveNegativeEnvision2022} & CVPR22 & ResNet12 & 68.11±0.81 & 72.41±0.72 & 83.12±0.48 & 79.85±0.57 & 70.58±0.93 & 73.43±0.78 & 85.38±0.61 & 81.64±0.63 \\
          & RFDNet\cite{dengLearningRelativeFeature2023} & TMM22 & ResNet12 & 66.23±0.80 & 71.37±0.80 & 82.44±0.54 & 80.31±0.59 & 66.84±0.89 & 72.68±0.76 & 82.64±0.63 & 81.95±0.72 \\
    \bottomrule
    \end{tabular}%

    }
  \label{tab:openset}%
  \vspace{-3mm}
\end{table*}%

\vspace{-3mm}
\subsection{Few-shot Learning with Varying Classes} 
\subsubsection{Datasets and Settings}

In order to compare the IFSL approaches in a fair manner,  We select some representative IFSL methods and compare them with the same implementation requirements and configurations.  IFSL experiments are commonly conducted on three datasets, CIFAR-100\cite{CIFAR100}, \textit{mini}ImageNet\cite{MatchingNet} and CUB-200\cite{CUB}. Apart from \textit{mini}Imagenet which has been described, we describe CIFAR-100 and CUB-200 as follows.
\begin{itemize}
\item CIFAR-100 \cite{CIFAR100} has 100 categories and each category contains 500 training instances and 100 test instances. 
\item CUB-200 \cite{CUB} contains 200 categories of bird images, with 5994 images for training and 5794 images for testing. 
\end{itemize}
In terms of incremental learning settings, CIFAR-100 and \textit{mini}Imagenet take 60 categories as base classes, and the remaining 40 categories are continually learned in a $5$-way $5$-shot manner, for a total of 8 sessions. CUB-200 uses the first 100 categories as base classes and the remaining 100 categories as a $10$-way $5$-shot input to the model.
\subsubsection{Metrics}
For comprehensive performance comparisons, four commonly used metrics are applied to measure the performance of methods, including Start Accuracy (SA), End Accuracy (EA), Average Accuracy (AA), and Performance Drop (PD). SA is the accuracy of the first sessions $\text{Acc}_1$. EA is the accuracy of the last sessions $\text{Acc}_T$, where $T$ is the number of all sessions.
AA is accuracy across all incremental sessions:
\begin{equation}
\begin{aligned}
     \text{AA}=\frac{1}{\textit{T}} \sum_{t=1}^{T} \text{Acc}_{t} 
\end{aligned}
\end{equation}
PD measures the absolute accuracy drops in the last session:
\begin{equation}
    \text{PD}=\text{Acc}_{T}-\text{Acc}_1
\end{equation}

\subsubsection{Performance Comparison and Analysis}

In Table \ref{table-IFSL1}, we summarize the performance of selected IFSL approaches evaluated on
the commonly used benchmarks, CIFAR-100, \textit{mini}ImageNet, and CUB-200. 
All the results come from the original papers. Due to significant differences in the initial accuracy of these methods, the direct comparison of AA does not reflect the true capability of the model. We evaluate these methods mainly considering the PD after incremental learning.

Among the three types of methods, model-based methods are inconsistently more stable and superior than the other two types of methods. It is worth noting that, CPE-CLIP\cite{CPECLIP}, with the language-image pre-training model, performs best in all methods. It reveals that the language-image pairs with abundant semantic information effectively guide the learning of incremental new tasks. In the meantime, CPE-CLIP, which applies the CLIP model \cite{CLIP} pre-trained on out-of-distribution datasets, is unfair to directly compare with other methods.
In the data-based methods, the UadCE with unlabeled samples almost outperforms other data-based methods with the least PD, especially in CIFAR-100 and CUB-200. In \textit{mini}Imagenet, the superiority of UadCE\cite{UadCE} is relatively weakened. In the model-based methods, apart from CPE-CLIP, BiDist\cite{BiDist} showcases good performance as well, which might be attributed to the attention fusion mechanism to leverage the discrepancy between base classes and novel classes. In the feature-based methods, TEEN\cite{TEEN} and EHS\cite{EHS} work relatively better than others. F2M\cite{F2M} as a representative of early feature-based method still maintains a strong competitiveness.

\begin{table*}[htbp]
	\centering
	\caption{The performance of selected IFSL methods on CIFAR-100, \textit{mini}Imagenet and CUB-100. The ``$\downarrow$'' after the distance measures indicates ``the smaller the better'', and the ``$\uparrow$'' after the similarity measures indicates ``the larger the better'' }
	\resizebox{1.925\columnwidth}{!}{
		\begin{tabular}{lccccccccccccccccc}
    \toprule
    \multirow{2}[1]{*}{Type} & \multirow{2}[1]{*}{Methods} & \multirow{2}[1]{*}{Venue} & \multicolumn{5}{c}{CIFAR-100}         & \multicolumn{5}{c}{\textit{mini}ImageNet}      & \multicolumn{5}{c}{CUB-200} \\
\cmidrule{4-18}          &       &       & Backbone & SA    & EA$\uparrow$   & AA$\uparrow$   & PD$\downarrow$   & Backbone & SA    & EA$\uparrow$   & AA$\uparrow$   & PD$\downarrow$   & Backbone & SA    & EA$\uparrow$   & AA$\uparrow$   & PD$\downarrow$ \\
    \midrule
    \multicolumn{1}{c}{\multirow{4}[2]{*}{\rotatebox{90}{Data Strategy}}}
            & SPPR\cite{SPPR}  & CVPR21 & ResNet18 & 76.68  & 48.12  & 60.84  & 28.56  & ResNet18 & 78.68  & 48.12  & 60.84  & 30.56  & ResNet18 & 68.68  & 37.33  & 49.32  & 31.35  \\
          & FeSSSS\cite{FeSSSS} & CVPRW21 & ResNet20 & 75.35  & 50.23  & 60.92  & 25.12  & ResNet18 & 81.50  & {58.87}  & 68.23  & 22.63  & ResNet18 & 79.60  & 52.98  & 62.85  & 26.62  \\
          & FSIL-GAN\cite{FSILGAN} & ACM MM22 & ResNet18 & 70.14  & 46.61  & 55.31  & 23.53  & ResNet18 & 69.87  & 46.14 & 56.39 & 23.73 & ResNet18 & 81.25 & 59.21 & 69.34 & 22.04 \\
          & ERDR\cite{ERDR}  & ECCV22 & ResNet20 & 74.40  & 50.14  & 60.77  & 24.26  & ResNet18 & 71.84  & 48.21  & 58.02  & 23.63  & ResNet18 & 75.90  & 53.39  & 61.52  & 22.51  \\
          & UadCE\cite{UadCE} & TNNLS23 & ResNet18 & 75.55  & 54.50  & 63.93  & {21.05}  & ResNet18 & 72.35  & 50.52  & 58.82  & 21.83  & ResNet18 & 75.17  & 60.72  & 65.70  & {14.45}  \\
    \midrule
    \multicolumn{1}{c}{\multirow{7}[2]{*}{\rotatebox{90}{Network Ensemble}}} 
          & CEC\cite{CEC}   & CVPR21 & ResNet20 & 73.07  & 49.14  & 59.53  & 23.93  & ResNet18 & 72.00  & 47.63  & 57.75  & 24.37  & ResNet18 & 75.85  & 52.28  & 61.33  & 23.57  \\
          & MetaFSCIL\cite{MetaFSCIL} & CVPR22 & ResNet20 & 74.50  & 49.97  & 60.79  & 24.53  & ResNet18 & 72.04  & 49.19  & 58.85  & 22.85  & ResNet18 & 75.90  & 52.64  & 61.92  & 23.26  \\
          & BiDist\cite{BiDist} & CVPR23 & ResNet20 & 79.45  & { 56.56}  & {66.55}  & 22.89  & ResNet18 & 74.65  & 53.39  & 62.06  & {21.26}  & ResNet18 & 79.12  & {63.81}  & {68.61}  & {15.31}  \\
          & MCNet\cite{MCNet} & TIP23   & ResNet20 & 73.30  & 50.72  & 60.39  & 22.58  & ResNet18 & 72.33  & 49.08  & 58.64  & 23.25  & ResNet18 & 77.57  & 59.08  & 65.57  & 18.49  \\
          & LIMIT\cite{LIMIT} & TPAMI23 & ResNet20 & 73.81  & 51.23  & 61.84  & 22.58  & ResNet18 & 72.32  & 49.19  & 59.06  & 23.13  & ResNet18 & 75.89  & 57.41  & 65.45  & 18.48  \\
          & CPE-CLIP$^\dagger$\cite{CPECLIP} & ICCVW23 & ViT-b\dag & 87.83  & 80.52  & 83.42  & 7.31  & ViT-b\dag & 90.23  & 82.77  & 86.13  & 7.46  & ViT-b\dag & 81.58  & 64.60  & 70.79  & 16.98  \\
          & M2SD\cite{M2SD}  & AAAI24 & -  & -     & -     & -     & -     & ResNet18 & 82.11  & 56.51  & {68.63}  & 25.60  & ResNet18 & {81.49}  & 60.96  & 68.14  & 20.53  \\
    \midrule
    \multicolumn{1}{c}{\multirow{11}[2]{*}{\rotatebox{90}{Feature Regularization}}}
          & TOPIC\cite{TOPIC} & CVPR20 & ResNet18 & 64.10  & 29.37  & 38.36  & 34.73  & ResNet18 & 61.31  & 24.42  & 35.68  & 36.89  & ResNet18 & 68.68  & 26.28  & 43.92  & 42.40  \\
          & F2M\cite{F2M}   & NeurIPS21 & ResNet18 & 64.71  & 44.67  & 53.65  & {20.04}  & ResNet18 & 67.28  & 44.65  & 54.89  & 22.63  & ResNet18 & 81.07  & 60.26  & 67.24  & 20.81  \\
          & DSN\cite{DSN}   & TPAMI22 & ResNet18 & 73.00  & 49.35  & 59.99  & 23.65  & ResNet18 & 68.95  & 45.89  & 54.83  & 23.06  & ResNet18 & 76.06  & 54.21  & 63.31  & 21.85  \\
          & CLOM\cite{CLOM}  & NeurIPS22 & ResNet20 & 74.20  & 50.25  & 60.56  & 23.95  & ResNet18 & 73.08  & 48.00  & 58.48  & 25.08  & ResNet18 & 79.57  & 59.58  & 67.17  & 19.99  \\
          & FACT\cite{FACT}  & CVPR22 & ResNet20 & 74.60  & 52.10  & 62.24  & 22.50  & ResNet18 & 72.56  & 50.49  & 60.70  & 22.07  & ResNet18 & 75.90  & 56.94  & 64.42  & 18.96  \\
          & NC-FSCIL\cite{NCFSCIL} & ICLR23 & ResNet12 & {82.52}  & {56.11}  & {67.50}  & 26.41  & ResNet12 & {84.02}  & {58.31}  & 67.82  & 25.71  & ResNet18 & 80.45  & 59.44  & 67.28  & 21.01  \\
          & WaRP\cite{WaRP}  & ICLR23 & ResNet20 & {80.31}  & 54.74  & 65.82  & 25.57  & ResNet18 & 72.99  & 50.65  & 59.69  & 22.34  & ResNet18 & 77.74  & 57.01  & 62.15  & 20.73  \\
          & TEEN\cite{TEEN}  & NeurIPS23 & ResNet20 & 74.92  & 52.64  & 63.10  & 22.28  & ResNet18 & 73.53  & 52.08  & 61.43  & {21.45}  & ResNet18 & 77.26  & 59.31  & 66.62  & 17.95  \\
          & SAVC\cite{SAVC}  & CVPR23 & ResNet20 & 78.77  & 53.12  & 63.63  & 25.65  & ResNet18 & 81.12  & 57.11  & 67.05  & 24.01  & ResNet18 & {81.85}  & {62.50}  & {69.35}  & 19.35  \\
          & EHS\cite{EHS}   & WACV24 & ResNet20 & \textcolor[rgb]{ .141,  .125,  .129}{71.27 } & 49.43  & 58.87  & 21.84  & ResNet18 & \textcolor[rgb]{ .141,  .125,  .129}{69.43 } & 47.67  & 56.51  & 21.76  & - & -     & -     & -     & - \\
          & MICS\cite{MICS}  & WACV24 & ResNet20 & 78.18  & 52.94  & 63.62  & 25.24  & ResNet18 & {84.40}  & 60.74  & {70.17}  & 23.66  & ResNet18 & 78.77  & 61.37  & 67.39  & 17.40  \\
    \bottomrule
    \end{tabular}%

	}
	\label{table-IFSL1}%
    \begin{tablenotes}
        \footnotesize
        \item {$^\dagger$} CPE-CLIP applies the pre-trained model from OpenAI CLIP and is not suitable for comparison with other methods.
    \end{tablenotes}
      \vspace{-3mm}
\end{table*}%

%% file: chapters/cross-domain/performance_CD.tex
\vspace{-3mm}
\subsection{Few-shot Learning with Varying Distributions}
\begin{table*}[htbp]
  \centering
  \caption{The CDFSL performance of the proposed methods on BSCD-FSL and mini-CUB benchmarks. $K$ means 5-way $K$-shot.}
	\resizebox{1.925\columnwidth}{!}{
    \begin{tabular}{cccccccccccccc}
    \toprule
    \multicolumn{1}{c}{Type} & \multicolumn{1}{c}{methods} & \multicolumn{1}{c}{Venue} & \multicolumn{1}{c}{Train Set} & \multicolumn{1}{c}{BackBone} & \multicolumn{1}{c}{$K$-shot} & ChestX & ISIC & EuroSAT & CropDisease & CUB & Cars & Places & Plantae \\
    \midrule
    \multicolumn{1}{c}{\multirow{12}{*}{\rotatebox{90}{Data Augmentation}}} & \multicolumn{1}{c}{\multirow{2}{*}{FDM\cite{c3_fu2021meta}}} & \multicolumn{1}{c}{\multirow{2}{*}{ACM MM21}} 
    & \multirow{2}{2cm}{\centering \textit{mini}ImageNet \\ Auxiliary Set} & \multicolumn{1}{c}{\multirow{2}{*}{ResNet10}} & 1  & -  & -  & -  & -  & 63.24±0.82 & 51.31±0.83 & 58.22±0.82 & 51.03±0.81 \\
       &    &    &    &    & 5  & -  & -  & -  & -  & 79.46±0.63 & 66.52±0.70 & 78.92±0.63 & 69.22±0.65 \\
\cmidrule{2-14}       & \multicolumn{1}{c}{\multirow{3}{*}{NSAE \cite{c39_liang2021boosting}}} & \multicolumn{1}{c}{\multirow{3}{*}{ICCV21}} & \multicolumn{1}{c}{\multirow{3}{*}{\textit{mini}ImageNet}} & \multicolumn{1}{c}{\multirow{3}{*}{ResNet10}} & 5  & 27.30±0.42 & {55.27±0.62} & 84.33±0.55 & 93.31±0.42 & 71.92±0.77 & 58.30±0.75 & 73.17±0.72 & 62.15±0.77 \\
       &    &    &    &    & 20 & 35.70±0.47 & 67.28±0.61 & 92.34±0.35 & 98.33±0.18 & 88.09±0.48 & 82.32±0.50 & 82.50±0.59 & 77.40±0.65 \\
       &    &    &    &    & 50 & 38.52±0.71 & 72.90±0.55 & 95.00±0.26 & 99.29±0.14 & 91.00±0.79 & -  & 85.92±0.56 & 83.63±0.60 \\
\cmidrule{2-14}       & \multicolumn{1}{c}{\multirow{3}{*}{TACDFSL \cite{c7_zhang2022tacdfsl}}} & \multicolumn{1}{c}{\multirow{3}{*}{Symmetry22}} & \multicolumn{1}{c}{\multirow{3}{*}{\textit{mini}ImageNet}} & \multicolumn{1}{c}{\multirow{3}{*}{WideResNet}} & 5  & 25.32±0.48 & 45.39±0.67 & 85.19±0.67 & 93.42±0.55 & -  & -  & -  & - \\
       &    &    &    &    & 20 & 29.17±0.52 & 53.15±0.59 & 87.87±0.49 & 95.49±0.39 & -  & -  & -  & - \\
       &    &    &    &    & 50 & 31.75±0.51 & 56.68±0.58 & 89.07±0.43 & 95.88±0.35 & -  & -  & -  & - \\
\cmidrule{2-14}       & \multicolumn{1}{c}{\multirow{2}{*}{StyleAdv \cite{c9_fu2023styleadv}}} & \multicolumn{1}{c}{\multirow{2}{*}{CVPR23}} & \multicolumn{1}{c}{\multirow{2}{*}{\textit{mini}ImageNet}} & \multicolumn{1}{c}{\multirow{2}{*}{ResNet10}} & 1  & 22.64±0.35 & 33.96±0.57 & 70.94±0.82 & 74.13±0.78 & 48.49±0.72 & 34.64±0.57 & 58.58±0.83 & 41.13±0.67 \\
       &    &    &    &    & 5  & 26.07±0.37 & 45.77±0.51 & 86.58±0.54 & 93.65±0.39 & 68.72±0.67 & 50.13±0.68 & 77.73±0.62 & 61.52±0.68 \\
    \midrule
    \multicolumn{1}{c}{\multirow{10}{*}{\rotatebox{90}{Parameter Optimization}}} & \multicolumn{1}{c}{\multirow{2}{*}{ME-D2N \cite{c41_fu2022me}}} & \multicolumn{1}{c}{\multirow{2}{*}{ACM MM22}} & \multicolumn{1}{c}{\multirow{2}{2cm}{\centering \textit{mini}ImageNet\newline{}target data}} & \multicolumn{1}{c}{\multirow{2}{*}{ResNet10}} & 1  & -  & -  & -  & -  & 65.05±0.83 & 49.53±0.79 & 60.36±0.86 & 52.89±0.83 \\
       &    &    &    &    & 5  & -  & -  & -  & -  & 83.17±0.56 & 69.17±0.68 & 80.45±0.62 & 72.87±0.67 \\
\cmidrule{2-14}       & \multicolumn{1}{c}{\multirow{2}{*}{ReFine \cite{c30_oh2022refine}}} & \multicolumn{1}{c}{\multirow{2}{*}{ICKM22}} & \multicolumn{1}{c}{\multirow{2}{*}{\textit{mini}ImageNet}} & \multicolumn{1}{c}{\multirow{2}{*}{ResNet10}} & 1  & 22.48±0.41 & 35.30±0.59 & 64.14±0.82 & 68.93±0.84 & -  & -  & -  & - \\
       &    &    &    &    & 5  & 26.76±0.42 & 51.68±0.63 & 82.36±0.57 & 90.75±0.49 & -  & -  & -  & - \\
\cmidrule{2-14}       & \multicolumn{1}{c}{\multirow{5}{*}{SWP \cite{c56_ji2024soft}}} & \multicolumn{1}{c}{\multirow{5}{*}{TMM24}} & \multicolumn{1}{c}{\multirow{2}{2cm}{\centering \textit{mini}ImageNet\\ target data}} & \multicolumn{1}{c}{\multirow{2}{*}{ResNet10}} & 1  & {24.16±0.43} & {36.07±0.66} & 77.72±0.76 & 91.96±0.64 & 40.69±0.81 & 38.40±0.72 & 47.17±0.84 & 53.64±0.69 \\
       &    &    &    &    & 5  & {31.30±0.42} & 51.36±0.63 & 93.52±0.32 & 97.84±0.42 & 59.00±0.81 & 60.12±0.71 & 68.20±0.80 & 76.08±0.61 \\
\cmidrule{4-14}       &    &    & \multicolumn{1}{c}{\multirow{2}{2cm}{\centering  \textit{tierd}ImageNet \\ target data}} & \multicolumn{1}{c}{\multirow{2}{*}{ResNet18}} & \multicolumn{1}{c}{\multirow{2}{*}{5}}  & \multicolumn{1}{c}{\multirow{2}{*}{30.26±0.50}} & \multicolumn{1}{c}{\multirow{2}{*}{52.00±0.66}} & \multicolumn{1}{c}{\multirow{2}{*}{91.45±0.40}} & \multicolumn{1}{c}{\multirow{2}{*}{97.48±0.21}} & \multicolumn{1}{c}{\multirow{2}{*}{-}}  & \multicolumn{1}{c}{\multirow{2}{*}{-}}  & \multicolumn{1}{c}{\multirow{2}{*}{-}}  & \multicolumn{1}{c}{\multirow{2}{*}{-}} \\

\\
    \midrule
    \multicolumn{1}{c}{\multirow{19}{*}{\rotatebox{90}{Feature Processing}}} & \multicolumn{1}{c}{HVM \cite{c13_du2021hierarchical}} & \multicolumn{1}{c}{ICLR21} & \multicolumn{1}{c}{\textit{mini}ImageNet} & \multicolumn{1}{c}{ResNet10} & 5  & 27.15±0.45 & 42.05±0.34 & 74.88±0.45 & 87.65±0.35 & -  & -  & -  & - \\
\cmidrule{2-14}       & \multicolumn{1}{c}{\multirow{3}{*}{Confess \cite{c16_das2021confess}}} & \multicolumn{1}{c}{\multirow{3}{*}{ICLR22}} & \multicolumn{1}{c}{\multirow{3}{*}{\textit{mini}ImageNet}} & \multicolumn{1}{c}{\multirow{3}{*}{ResNet10}} & 5  & 27.09±0.24 & 48.85±0.29 & 84.65±0.38 & 88.88±0.51 & -  & -  & -  & - \\
       &    &    &    &    & 20 & 33.57±0.31 & 60.31±0.33 & 90.40±0.24 & 95.34±0.48 & -  & -  & -  & - \\
       &    &    &    &    & 50 & 39.02±0.12 & 65.34±0.45 & 92.66±0.36 & 97.56±0.43 & -  & -  & -  & - \\
\cmidrule{2-14}       & \multicolumn{1}{c}{\multirow{2}{*}{RDC-FT \cite{c23_li2022ranking}}} & \multicolumn{1}{c}{\multirow{2}{*}{CVPR22}} & \multicolumn{1}{c}{\multirow{2}{*}{\textit{mini}ImageNet}} & \multicolumn{1}{c}{\multirow{2}{*}{ResNet10}} & 1  & 22.27±0.2 & 35.84±0.4 & 71.57±0.5 & 86.33±0.5 & 51.20±0.5 & 39.13±0.5 & 61.50±0.6 & 44.33±0.6 \\
       &    &    &    &    & 5  & 25.48±0.2 & 49.06±0.3 & 84.67±0.3 & 93.55±0.3 & 67.77±0.4 & 53.75±0.5 & 74.65±0.4 & 60.63±0.4 \\
\cmidrule{2-14}       & \multicolumn{1}{c}{\multirow{4}{*}{VDB \cite{c27_yazdanpanah2022visual}}} & \multicolumn{1}{c}{\multirow{4}{*}{CVPR22}} & \multicolumn{1}{c}{\multirow{4}{*}{\textit{mini}ImageNet}} & \multicolumn{1}{c}{\multirow{4}{*}{ResNet10}} & 1  & {22.99±0.44} & 35.32±0.65 & 63.60±0.87 & 71.98±0.82 & -  & -  & -  & - \\
       &    &    &    &    & 5  & 26.62±0.45 & 48.72±0.65 & 82.06±0.63 & 90.77±0.49 & -  & -  & -  & - \\
       &    &    &    &    & 20 & 31.87±0.44 & 59.09±0.59 & 89.42±0.45 & 96.36±0.27 & -  & -  & -  & - \\
       &    &    &    &    & 50 & 35.55±0.45 & 64.02±0.58 & 92.24±0.35 & 97.89±0.19 & -  & -  & -  & - \\
\cmidrule{2-14}       & \multicolumn{1}{c}{\multirow{2}{*}{ProD \cite{c43_ma2023prod}}} & \multicolumn{1}{c}{\multirow{2}{*}{CVPR23}} & \multicolumn{1}{c}{\multirow{2}{2cm}{\centering \textit{mini}ImageNet\\ target data}} & \multicolumn{1}{c}{\multirow{2}{*}{ResNet10}} & 1  & -  & -  & -  & -  & 53.97±0.71 & 38.02±0.63 & 42.86±0.59 & 53.92±0.72 \\
       &    &    &    &    & 5  & -  & -  & -  & -  & 79.19±0.59 & 59.49±0.68 & 65.82±0.65 & 75.00±0.72 \\
\cmidrule{2-14}       & \multicolumn{1}{c}{\multirow{3}{*}{Dara \cite{C50_zhao2023dual}}} & \multicolumn{1}{c}{\multirow{3}{*}{ITPAMI23}} & \multicolumn{1}{c}{\multirow{3}{2cm}{\centering \textit{mini}ImageNet\\ target data}} & \multicolumn{1}{c}{\multirow{3}{*}{ResNet10}} & 1  & 22.92±0.40 & 36.42±0.64 & 67.42±0.80 & 80.74±0.76 & -  & -  & -  & - \\
 &    &    &    &    & 5  & 27.54±0.42 & {56.28±0.66} & 85.84±0.54 & 95.32±0.34 & -  & -  & -  & - \\
 &    &    &    &    & 20 & 35.95±0.47 & 67.36±0.57 & 67.42±0.80 & 98.58±0.15 & -  & -  & -  & - \\
    \bottomrule
    \end{tabular}%
  }
  \label{CD_Table1}%
    \vspace{-3mm}
\end{table*}%

\subsubsection{Datasets and Settings}
CDFSL has various experimental setups including different datasets, backbones and configurations, we compare some representative CDFSL methods in a unified and fair manner. CDFSL experiences are facilitated by the availability of the several datasets including \textit{mini}ImageNet \cite{MatchingNet}, ChestX \cite{c58_wang2017chestx}, ISIC \cite{c59_tschandl2018ham10000,c60_codella2019skin}, EuroSAT \cite{c61_helber2019eurosat}, CropDiseases \cite{c62_mohanty2016using}, CUB \cite{CUB}, Cars \cite{c64_krause20133d}, Places \cite{c65_zhou2017places}, Plantae \cite{c66_van2018inaturalist}, Omniglot \cite{c67_lake2011one}, Aircraft \cite{c68_maji2013fine}, Textures \cite{c69_cimpoi2014describing}, Quick Draw \cite{c70_fernandez2019quick}, Fungi \cite{c71_schroeder2018fgvcx}, VGG Flower \cite{c72_nilsback2008automated}, Traffic Signs \cite{c73_houben2013detection} and MSCOCO \cite{c74_lin2014microsoft}. Among them, there are three mainly benchmarks of the CDFSL problem, including BSCD-FSL \cite{c75_guo2020broader}, mini-CUB \cite{c28_tseng2020cross} and Meta-Dataset  \cite{c76_triantafillou2019meta}. 

BSCD-FSL (Broader Study of Cross-Domain Few-Shot Learning) \cite{c75_guo2020broader} benchmark includes five image datasets from a diverse assortment of image acquisition methods.

\begin{itemize}
\item ChestX \cite{c58_wang2017chestx} comprises 112,120 frontal-view X-ray images of 30,805 unique patients with the text-mined fourteen common diseases.
\item ISIC \cite{c59_tschandl2018ham10000,c60_codella2019skin} contains 10015 dermoscopic lesion images from 7 categories for training, 193 images for evaluation, and 1512 images for testing.
\item EuroSAT \cite{c61_helber2019eurosat} consists of 10 classes with in total 27,000 labeled and geo-referenced images.
\item CropDiseases \cite{c62_mohanty2016using} contains about 87K RGB images of healthy and diseased crop leaves which are categorized into 38 different classes.
\end{itemize}

mini-CUB \cite{c28_tseng2020cross} includes five datasets. 
In this benchmark, \textit{mini}ImageNet is always regarded as the source domain and other datasets are regarded as the target domains. 
\begin{itemize}
\item CUB \cite{CUB} contains 200 different categories of bird images, with a total of 11788 images.
\item Cars \cite{c64_krause20133d} consists of 196 classes of cars with a total of 16,185 images.
\item Places \cite{c65_zhou2017places} is finalized with over 10 million labeled exemplars from 434 place categories.
\item Plantae \cite{c66_van2018inaturalist} has 2101 categories and 196613 images.
\end{itemize}

Meta-dataset \cite{c76_triantafillou2019meta}:  Meta-dataset benchmark is a large few-shot learning benchmark and consists of multiple datasets of different data distributions, consisting of 10 datasets from diverse domains. It does not restrict few-shot tasks to have fixed ways and shots, thus representing a more realistic scenario.

The standard evaluation metric used in CDFSL is prediction accuracy and there are various settings, including 5-way 1-shot, 5-way 5-shot, 5-way 20-shot, and 5-way 50-shot. Moreover, backbones mainly include Resnet10 and Resnet18.

\subsubsection{Performance Comparison and Analysis} In TABLE \ref{CD_Table1}, we summarize the performance of the representative CDFSL methods evaluated on the BSCD-FSL and mini-CUB benchmarks. All the results come from the original papers. 

Among the three types of methods, parameter optimization methods achieve optimal results on most datasets. The reason behind this may be that the introduction of additional parameters specific to the target domain allows for increased generalization of the model. The results show that the performance decreases when the distance between the source domain and the target domain increases. Data-based methods for CDFSL are relatively simple as they mostly enhance the generalization of the model by augmenting the data or introducing supplementary information. Among data-based methods, StyleAcdv \cite{c9_fu2023styleadv} achieves optimal results due to the stronger generalization of the adversarial samples generated by the introduction of stylistic noise. Parameter-based methods are more stable. For example, using ResNet10 as the backbone, the results of SWP \cite{c56_ji2024soft} on BSCD-FSL and mini-CUB (5-way 5-shot) demonstrate this trend, with scores of 31.30\% on ChestX, 93.52\% on EuroSAT, 97.84\% on CropDisease and 76.08\% on Plantae. Moreover, in the feature-based methods, the results of ProD \cite{c43_ma2023prod} and Dara \cite{C50_zhao2023dual} show that introducing target domain information in the training process can effectively improve cross-domain classification results.

%% file: chapters/Open_Challenges_and_Future_Detections.tex
\section{Open Challenges and Future Detections}

Despite significant progress in OFSL, it continues to present unique challenges that require
attention. As such, we outline several promising research directions for the future.

\textit{Few-Shot Learning with Different Concepts.} Conventional FSL models are trained with an assumption of a single problem, i.e., each instance can only encapsulate one visual concept even though in a multi-class. In the open world, such as fashion recognition \cite{FashionImage}, bioinformatics \cite{BarutcuogluST06}, and drug discovery \cite{HeiderSCH13}, each instance in few-shot scenarios may be interpreted by different concepts. To address this challenge, recent advancements in
multi-label learning have been adopted in FSL to combat multi-label issues\cite{LaSO,KGGR,BCR}. For instance, LaSO \cite{LaSO} employs a data augmentation strategy that generates synthetic feature vectors through label-set operations. KGGR \cite{KGGR} adopts a GCN where labels are modeled as nodes and statistical label co-occurrences are modeled as edges to exploit the label dependencies. The above methods assume that the importance degree of different instances for a certain label is the same and that of different labels in an instance is also the same. BCR\cite{BCR} explores diverse important degrees and effectively leverages the underlying label correlations, along with the varying importance from both instance-to-label and label-to-instance perspectives.

\textit{Few-Shot Learning with Adversarial Attacks.} Conventional FSL models are crafted based on original clean examples without imperceptible perturbations\cite{GoodfellowSS14,ChakrabortyADCM21}. In the open world, these models are vulnerable to adversarial examples, which easily misclassify these
adversarial examples\cite{SchmidtSTTM18}. Therefore, it is crucial to learn of robust few-shot models in the open world that can effectively defend against adversarial
attacks \cite{GoodfellowSS14}. Recent works of few-shot learning methods defensive against adversarial attacks are proposed \cite{ShafahiSZGSJG20,GoldblumFG20,LiWZQHGL23}. ADML\cite {ShafahiSZGSJG20} finds that retraining earlier layers of the network during fine-tuning impairs the robustness of the network, while only retraining later layers can largely preserve robustness. AQ \cite{GoldblumFG20} adopts projected gradient descent attack (PGD) \cite{MadryMSTV18} by crafting adversarial perturbations and minimizes the losses of perturbed query data. DFSL \cite{LiWZQHGL23}  performs episode-based adversarial training at the task level. In the meantime, DFSL assumes distribution consistency between the clean and adversarial examples and enforces the distribution consistency between clean and adversarial examples from the feature-wise or prediction-wise perspectives within each task.

\textit{Few-Shot Learning with Imbalanced Classes.}  Conventional FSL models are trained with an assumption of the same number of data in each class. In the open world, it is unrealistic in aerial scene \cite{GuanLSFSW20} when the number of samples can vary and follow a wide range of imbalance distributions \cite{BudaMM18}. To address this challenge, recent advancements adopt random-shot tasks with randomly selected classes and samples \cite{GuanLSFSW20}. RF-MML implements random episodic training on the majority
of classes to learn generalized knowledge across data distribution and then fine-tunes all classes to further
improve the performance for multi-category classification. Bayesian Task-Adaptive Meta-Learning \cite{Lee2020Learning} can adaptively adjust the parameters that guide the scale of the meta-learner and the task-specific learner for each task and class.
CIFSL\cite{OchalPVSW23} formalizes the class imbalanced FSL setting at three levels of imbalance:  task-level, dataset-level, and combined imbalance. Then it adapts rebalancing methods such as random oversampling and various reweighting losses to the CIFSL
setting. Adaptive Manifold \cite{LazarouAS24} hypothesizes that class centroid approaches will benefit from exploiting the data manifold to obtain more representative centroids and iteratively updates both the class centroids and the manifold-specific parameters with label propagation.